\documentclass[11pt,a4paper]{article}
\usepackage[hyperref]{emnlp2018}
\usepackage{times}
\usepackage{latexsym}

\usepackage{url}

\aclfinalcopy 

\usepackage{hyperref}
\usepackage{latexsym}
\usepackage{tabularx}
\usepackage{graphicx}
\usepackage{float}
\usepackage{enumerate}
\usepackage{amssymb,amsmath}
\usepackage{color}
\usepackage{bm}
\usepackage{makecell}
\usepackage{multi row}
\usepackage{array}

\newcommand\Tstrut{\rule{0pt}{2.6ex}}         % = `top' strut
\newcommand\Bstrut{\rule[-1.0ex]{0pt}{0pt}}   % = `bottom' strut
\newcommand{\thinline}{\Xhline{1.0\arrayrulewidth}}
\newcommand{\thickline}{\Xhline{2.5\arrayrulewidth}}
\newcommand{\tsep}	{\Bstrut \\ \thinline}
\newcommand{\ttop}{\thickline}
\newcommand{\tbottom}{\Bstrut \\ \thickline}

\newcommand{\xhdr}[1]{\vspace{1.7mm}\noindent{{\bf #1.}}}

\newcommand\whethermath[1]{\ifmmode{#1}\else{$#1$}\fi}
\newcommand\uprm[1]{\whethermath{^{\hbox{\scriptsize #1}}}} 

\newcommand{\alns}[1] {
	\begin{align*} #1 \end{align*}
}
\newcommand{\relu} {\text{ReLU}}
\newcommand{\hf} { \overrightarrow{h} }
\newcommand{\hb} { \overleftarrow{h} }
\newcommand{\mlp}{ \texttt{NN}}
\newcommand{\loss} { \mathcal{L} }
\newcommand{\lsup} { \loss_\text{sup} }
\newcommand{\lunsup} { \loss_\text{CVT} }
\newcommand{\mlph}[1] { \relu(W_{\text{head}}#1 + b_{\text{head}}) }
\newcommand{\mlpd}[1] { \relu(W_{\text{dep}}#1 + b_{\text{dep}}) }
\newcommand{\model}[1]{ p_\theta(y|#1) }
\newcommand{\modelj}[1]{ p^j_\theta(y|#1) }
\newcommand{\ldata} {\mathcal{D}_l}
\newcommand{\udata} {\mathcal{D}_{ul}}
\newcommand{\D} {D}

\title{Semi-Supervised Sequence Modeling with Cross-View Training}

 \author{Kevin Clark$^1$ \hspace{3mm} Minh-Thang Luong$^2$ \hspace{3mm} Christopher D. Manning$^1$ \hspace{3mm} Quoc V. Le$^2$\\
   $^1$Computer Science Department, Stanford University \hspace{6mm} $^2$Google Brain \\
   \resizebox{0.99\textwidth}{!}{\tt kevclark@cs.stanford.edu, thangluong@google.com, manning@cs.stanford.edu, qvl@google.com} \\
 }

\date{}

\begin{document}

\maketitle

%-------------------------------------------------------
%                       Abstract
%-------------------------------------------------------

\begin{abstract}
Unsupervised representation learning algorithms such as word2vec and ELMo improve the accuracy of many supervised NLP models, mainly because they can take advantage of large amounts of unlabeled text.
However, the supervised models only learn from task-specific labeled data during the main training phase.
We therefore propose Cross-View Training (CVT), a semi-supervised learning algorithm that improves the representations of a Bi-LSTM sentence encoder using a mix of labeled and unlabeled data.
On labeled examples, standard supervised learning is used.
On unlabeled examples, CVT teaches auxiliary prediction modules that see restricted views of the input (e.g., only part of a sentence) to match the predictions of the full model seeing the whole input.
Since the auxiliary modules and the full model share intermediate representations, this in turn improves the full model. 
Moreover, we show that CVT is particularly effective when combined with multi-task learning.
We evaluate CVT on five sequence tagging tasks, machine translation, and dependency parsing, achieving state-of-the-art results.\footnote{Code is available at \url{https://github.com/tensorflow/models/tree/master/research/cvt_text}}
\end{abstract}

%-------------------------------------------------------
%                       Intro
%-------------------------------------------------------

\section{Introduction}

Deep learning models work best when trained on large amounts of labeled data.
However, acquiring labels is costly, motivating the need for effective semi-supervised learning techniques that leverage unlabeled examples.
A widely successful semi-supervised learning strategy for neural NLP is pre-training word vectors \citep{Mikolov2013DistributedRO}.
More recent work trains a Bi-LSTM sentence encoder to do language modeling and then incorporates its context-sensitive 
representations into supervised models \citep{dai2015semi,peters2018deep}. 
Such pre-training methods perform unsupervised representation learning on a large corpus of unlabeled data followed by supervised training.

A key disadvantage of pre-training is that the first representation learning phase does not take advantage of labeled data -- the model attempts to learn generally effective representations rather than ones that are targeted towards a particular task.
Older semi-supervised learning algorithms like self-training do not suffer from this problem because they continually learn about a task on a mix of labeled and unlabeled data.
Self-training has historically been effective for NLP \citep{yarowsky1995unsupervised,mcclosky2006effective}, but is less commonly used with neural models.
This paper presents Cross-View Training (CVT), a new self-training algorithm that works well for neural sequence models.

In self-training, the model learns as normal on labeled examples.
On unlabeled examples, the model acts as both a “teacher” that makes predictions about the examples and a “student” that is trained on those predictions. 
Although this process has shown value for some tasks, it is somewhat tautological: the model already produces the predictions it is being trained on. 
Recent research on computer vision addresses this by adding noise to the student's input, training the model so it is robust to input perturbations \citep{sajjadi2016regularization, wei2018improving}.
However, applying noise is difficult for discrete inputs like text.

As a solution, we take inspiration from multi-view learning \citep{blum1998combining, Xu2013ASO} and train the model to produce consistent predictions across different {\it views} of the input. 
Instead of only training the full model as a student, CVT adds auxiliary prediction modules -- neural networks that transform vector representations into predictions -- to the model and also trains them as students.
The input to each student prediction module is a subset of the model's intermediate representations corresponding to a restricted view of the input example.
For example, one auxiliary prediction module for sequence tagging is attached to only the ``forward" LSTM in the model's first Bi-LSTM layer, so it makes predictions without seeing any tokens to the right of the current one.

CVT works by improving the model's representation learning. 
The auxiliary prediction modules can learn from the full model's predictions because the full model has a better, unrestricted view of the input.
As the auxiliary modules learn to make accurate predictions despite their restricted views of the input, they improve the quality of the representations they are built on top of. 
This in turn improves the full model, which uses the same shared representations.
In short, our method combines the idea of representation learning on unlabeled data with classic self-training.

CVT can be applied to a variety of tasks and neural architectures, but we focus on sequence modeling tasks where the prediction modules are attached to a shared Bi-LSTM encoder. We propose auxiliary prediction modules that work well for sequence taggers, graph-based dependency parsers, and sequence-to-sequence models.
We evaluate our approach on English dependency parsing, combinatory categorial grammar supertagging, named entity recognition, part-of-speech tagging, and text chunking, as well as English to Vietnamese machine translation.
CVT improves over previously published results on all these tasks.
Furthermore, CVT can easily and effectively be combined with multi-task learning: we just add additional prediction modules for the different tasks on top of the shared Bi-LSTM encoder. 
Training a unified model to jointly perform all of the tasks except machine translation improves results (outperforming a multi-task ELMo model) while decreasing the total training time.

\begin{figure}[t]
\includegraphics[width=0.5\textwidth]{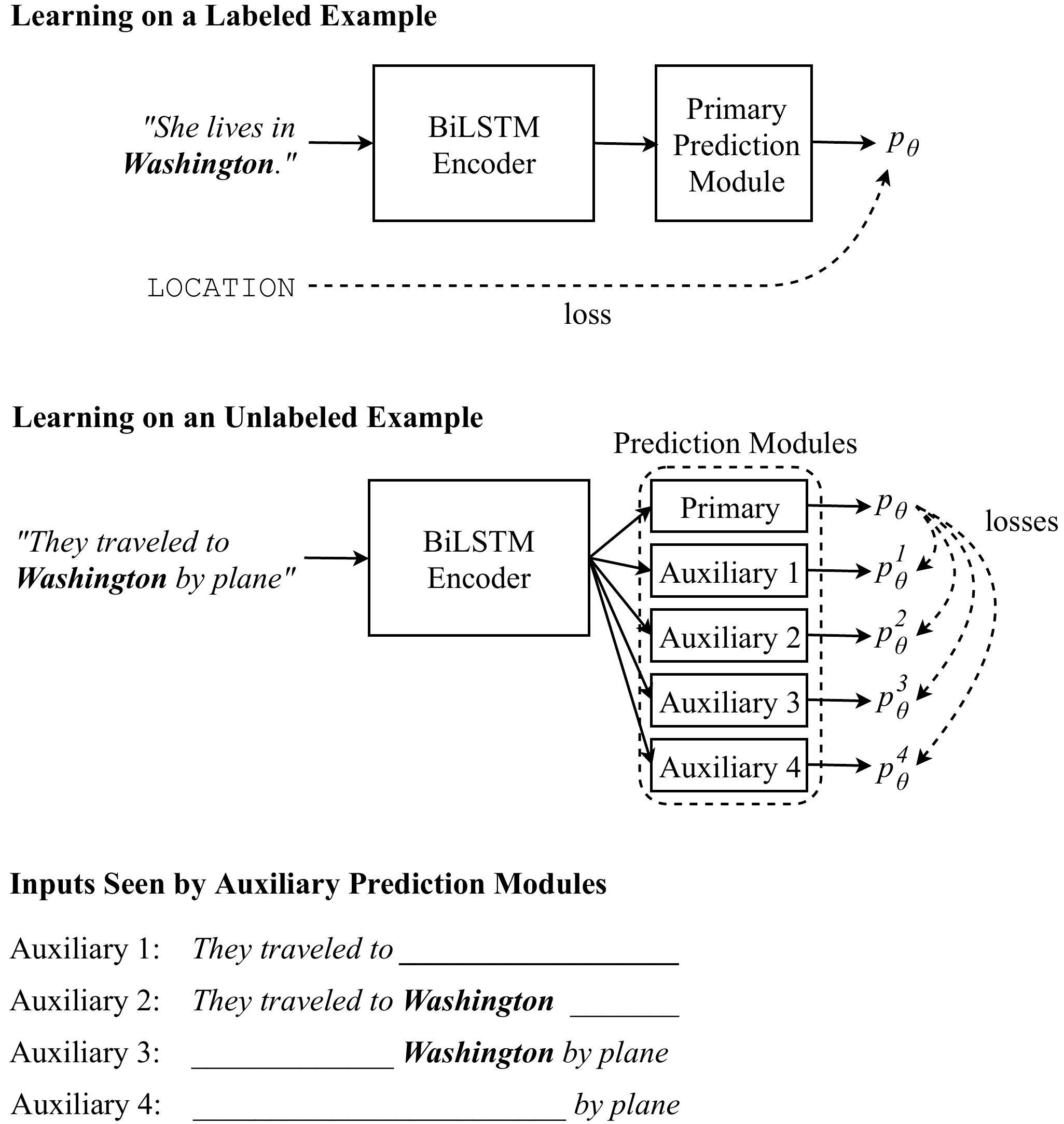}
\caption{An overview of Cross-View Training.
The model is trained with standard supervised learning on labeled examples.
On unlabeled examples, auxiliary prediction modules with different views of the input are trained to agree with the primary prediction module. 
This particular example shows CVT applied to named entity recognition. From the labeled example, the model can learn that ``Washington" usually refers to a location. 
Then, on unlabeled data, auxiliary prediction modules are trained to reach the same prediction without seeing some of the input. In doing so, they improve the contextual representations produced by the model, for example, learning that ``traveled to" is usually followed by a location.
}
\label{fig:overview}
\end{figure}

%-------------------------------------------------------
%                       Method
%-------------------------------------------------------

\section{Cross-View Training}
We first present Cross-View Training and describe how it can be combined effectively with multi-task learning. See Figure~\ref{fig:overview} for an overview of the training method.

\subsection{Method}
Let $\ldata = \{(x_1, y_1), (x_2, y_2), ..., (x_N, y_N)\}$ represent a labeled dataset and  $\udata = \{x_1, x_2, ..., x_M\}$ represent an unlabeled dataset
We use $\model{x_i}$ to denote the output distribution over classes produced by the model with parameters $\theta$ on input $x_i$.
During CVT, the model alternates learning on a minibatch of labeled examples and learning on a minibatch of unlabeled examples. 
For labeled examples, CVT uses standard cross-entropy loss:
\[
	\lsup(\theta) = \frac{1}{|\ldata|} \sum_{x_i, y_i \in \ldata} CE(y_i, \model{x_i})
\]

CVT adds $k$ auxiliary prediction modules to the model, which are used when learning on unlabeled examples.
A prediction module is usually a small neural network (e.g., a hidden layer followed by a softmax layer). 
Each one takes as input an intermediate representation $h^j(x_i)$ produced by the model (e.g., the outputs of one of the LSTMs in a Bi-LSTM model).
It outputs a distribution over labels $\modelj{x_i}$.
Each $h^j$ is chosen such that it only uses a part of the input $x_i$; the particular choice can depend on the task and model architecture.
We propose variants for several tasks in Section~\ref{sec:models}. 
The auxiliary prediction modules are only used during training; the test-time prediction come from the primary prediction module that produces $p_\theta$.

On an unlabeled example, the model first produces soft targets $\model{x_i}$ by performing inference. 
CVT trains the auxiliary prediction modules to match the primary prediction module on the unlabeled data by minimizing
\begin{align*}
	\resizebox{0.49\textwidth}{!}{$\lunsup(\theta) = \frac{1}{|\udata|}
	 \sum_{x_i \in \udata} \sum_{j = 1}^k \D(\model{x_i}, \modelj{x_i})$} 
\end{align*}
where $D$ is a distance function between probability distributions  (we use KL divergence).
We hold the primary module's prediction $\model{x_i}$ fixed during training (i.e., we do not back-propagate through it) so the auxiliary modules learn to imitate the primary one, but not vice versa. CVT works by enhancing the model's representation learning. As the auxiliary modules train, the representations they take as input improve so they are useful for making predictions even when some of the model's inputs are not available.
This in turn improves the primary prediction module, which is built on top of the same shared representations.

We combine the supervised and CVT losses into the total loss, $\loss = \lsup + \lunsup$, and minimize it with stochastic gradient descent. 
In particular, we alternate minimizing $\lsup$ over a minibatch of labeled examples and minimizing $\lunsup$ over a minibatch of unlabeled examples.

For most neural networks, adding a few additional prediction modules is computationally cheap compared to the portion of the model building up representations (such as an RNN or CNN).
Therefore our method contributes little overhead to training time over other self-training approaches for most tasks. CVT does not change inference time or the number of parameters in the fully-trained model because the auxiliary prediction modules are only used during training. 

\subsection{Combining CVT with Multi-Task Learning}

CVT can easily be combined with multi-task learning by adding additional prediction modules for the other tasks on top of the shared Bi-LSTM encoder.
During supervised learning, we randomly select a task and then update $\lsup$ using a minibatch of labeled data for that task.
When learning on the unlabeled data, we optimize $\lunsup$ jointly across all tasks at once, first running inference with all the primary prediction modules and then learning from the predictions with all the auxiliary prediction modules. 
As before, the model alternates training on minibatches of labeled and unlabeled examples.

Examples labeled across many tasks are useful for multi-task systems to learn from, but most datasets are only labeled with one task. 
A benefit of multi-task CVT is that the model creates (artificial) all-tasks-labeled examples from unlabeled data. 
This significantly improves the model's data efficiency and training time. 
Since running prediction modules is computationally cheap, computing $\lunsup$ is not much slower for many tasks than it is for a single one.
However, we find the all-tasks-labeled examples substantially speed up model convergence.
For example, our model trained on six tasks takes about three times as long to converge as the average model trained on one task, a 50\% decrease in total training time.

% --------------------------------------------------
%                     Models
%---------------------------------------------------

\section{Cross-View Training Models}
\label{sec:models}

CVT relies on auxiliary prediction modules that have restricted views of the input. In this section, we describe specific constructions of the auxiliary prediction modules that are effective for sequence tagging, dependency parsing, and sequence-to-sequence learning.

\subsection{Bi-LSTM Sentence Encoder}
All of our models use a two-layer CNN-BiLSTM \citep{chiu2015named,ma2016end} sentence encoder. 
It takes as input a sequence of words $x_i = [x_i^1, x_i^2, ..., x_i^T]$.
First, each word is represented as the sum of an embedding vector and the output of a character-level Convolutional Neural Network, resulting in a sequence of vectors $v = [v^1, v^2, ..., v^T]$.
The encoder applies a two-layer bidirectional LSTM \citep{graves2005framewise} to these representations. The first layer runs a Long Short-Term Memory unit \citep{hochreiter1997long} in the forward direction (taking $v^t$ as input at each step $t$) and the backward direction (taking $v^{T - t + 1}$ at each step) to produce vector sequences $[\hf^1_1, \hf_1^2, ... \hf_1^T]$ and $[\hb_1^1, \hb_1^2, ... \hb_1^T]$. The output of the Bi-LSTM is the concatenation of these vectors:
$
h_1 = [\hf_1^1 \oplus \hb_1^1, ..., \hf_1^T \oplus \hb_1^T]. 
$
The second Bi-LSTM layer works the same, producing outputs $h_2$, except it takes $h_1$ as input instead of $v$.

\subsection{CVT for Sequence Tagging}
\label{sec:tag}
In sequence tagging, each token $x^t_i$ has a corresponding label $y^t_i$. 
The primary prediction module for sequence tagging produces a probability distribution over classes for the $t\uprm{th}$ label using a one-hidden-layer neural network applied to the corresponding encoder outputs:
\alns{
   p(y^t|x_i) &= \mlp(h_1^t \oplus h_2^t) \\
   &= \text{softmax}(U\cdot \relu(W (h_1^t \oplus h_2^t)) + b)
}

\begin{figure}[t]
\includegraphics[width=0.46\textwidth]{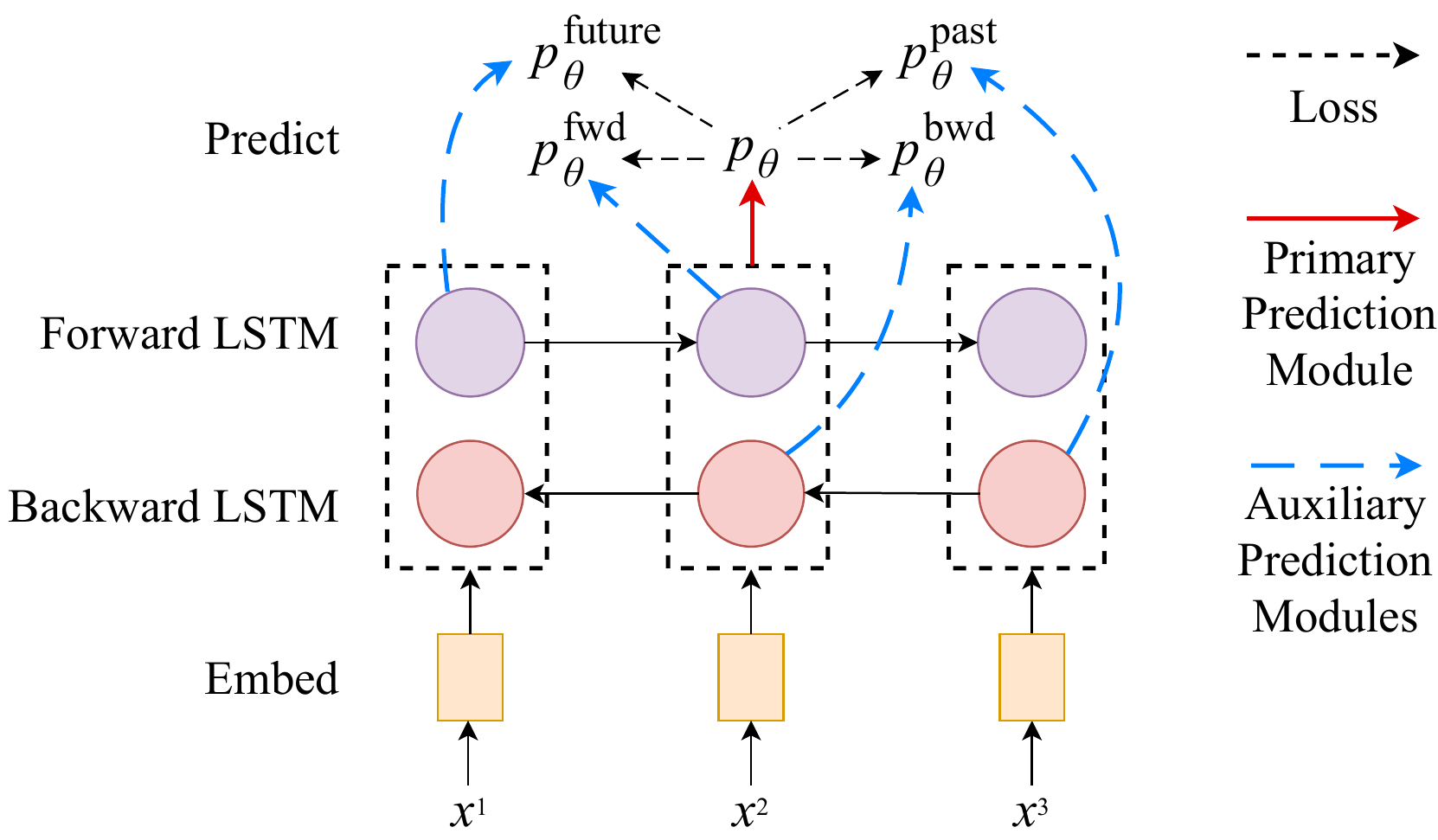}
\caption{Auxiliary prediction modules for sequence tagging models. Each one sees a restricted view of the input. For example, the ``forward" prediction module does not see any context to the right of the current token when predicting that token's label. For simplicity, we only show a one layer Bi-LSTM encoder and only show the model's predictions for a single time step.}
\label{fig:seq}
\end{figure}

The auxiliary prediction modules take $\hf_1(x_i)$ and $\hb_1(x_i)$, the outputs of the forward and backward LSTMs in the first\footnote{Modules taking inputs from the second Bi-LSTM layer would not have restricted views because information about the whole sentence gets propagated through the first layer.} Bi-LSTM layer, as inputs.
We add the following four auxiliary prediction modules to the model (see Figure~\ref{fig:seq}):
\begin{align*}
	p_\theta^{\text{fwd}}(y^t | x_i) &= \mlp^{\text{fwd}}(\hf_1^t(x_i)) \\ 
	p_\theta^{\text{bwd}}(y^t | x_i) &= \mlp^{\text{bwd}}(\hb_1^t(x_i))  \\
	p_\theta^{\text{future}}(y^t | x_i) &= \mlp^{\text{future}}(\hf_1^{t - 1}(x_i)) \\ 
	p_\theta^{\text{past}}(y^t | x_i) &= \mlp^{\text{past}}(\hb_1^{t + 1}(x_i))
\end{align*}
The ``forward" module makes each prediction without seeing the right context of the current token. 
The ``future" module makes each prediction without the right context or the current token itself.
Therefore it works like a neural language model that, instead of predicting which token comes next in the sequence, predicts which class of token comes next. The ``backward" and ``past" modules are analogous.

\subsection{CVT for Dependency Parsing}
\label{sec:dep}
In a dependency parse, words in a sentence are treated as nodes in a graph. 
Typed directed edges connect the words, forming a tree structure describing the syntactic structure of the sentence. 
In particular, each word $x^t_i$ in a sentence $x_i = x^1_i, ..., x^T_i$ receives exactly one in-going edge $(u, t, r)$ going from word $x_i^u$ (called the ``head'') to it (the ``dependent'') of type $r$ (the ``relation''). 
We use a graph-based dependency parser similar to the one from \citet{Dozat2017Deep}. 
This treats dependency parsing as a classification task where the goal is to predict which in-going edge $y_i^t = (u, t, r)$ connects to each word $x_i^t$.

First, the representations produced by the encoder for the candidate head and dependent are passed through separate hidden layers. A bilinear classifier applied to these representations produces a score for each candidate edge. Lastly, these scores are passed through a softmax layer to produce probabilities. Mathematically, the probability of an edge is given as:
\[
p_\theta((u, t, r)|x_i) \propto e^{s(h_{1}^u(x_i) \oplus h_{2}^u(x_i), h_{1}^t(x_i) \oplus h_{2}^t(x_i), r)}
\]
where $s$ is the scoring function: 
\alns{
s(z_1, z_2, r) = \phantom{.}&\mlph{z_1}(W_r + W) \\ &\mlpd{z_2}
}
The bilinear classifier uses a weight matrix $W_r$ specific to the candidate relation as well as a weight matrix $W$ shared across all relations. Note that unlike in most prior work, our dependency parser only takes words as inputs, not words and part-of-speech tags.

We add four auxiliary prediction modules to our model for cross-view training:
\begin{align*}
	&p_\theta^{\text{fwd-fwd}}((u, t, r) | x_i) \propto e^{s^{\text{fwd-fwd}}(\hf_1^u(x_i), \hf_1^t(x_i), r)} \\ 
	& p_\theta^{\text{fwd-bwd}}((u, t, r) | x_i) \propto e^{s^{\text{fwd-bwd}}(\hf_1^u(x_i), \hb_1^t(x_i), r)} \quad \\
	&p_\theta^{\text{bwd-fwd}}((u, t, r) | x_i) \propto e^{s^{\text{bwd-fwd}}(\hb_1^u(x_i), \hf_1^t(x_i), r)} \\ 
	&p_\theta^{\text{bwd-bwd}}((u, t, r) | x_i)\propto e^{s^{\text{bwd-bwd}}(\hb_1^u(x_i), \hb_1^t(x_i), r)}
\end{align*}
Each one has some missing context (not seeing either the preceding or following words) for the candidate head and candidate dependent. 

\subsection{CVT for Sequence-to-Sequence Learning}
\label{sec:seq2seq}
We use an encoder-decoder sequence-to-sequence model with attention \citep{sutskever2014sequence,bahdanau2014neural}. Each example consists of an input (source) sequence $x_i = x^1_i, ..., x^T_i$ and output (target) sequence  $y_i = y^1_i, ..., y^K_i$. The encoder's representations are passed into an LSTM decoder using a bilinear attention mechanism \citep{Luong2015EffectiveAT}. In particular, at each time step $t$ the decoder computes an attention distribution over source sequence hidden states as $\alpha_j \propto e^{h^j W_\alpha \bar{h}^t}$ where $\bar{h}^t$ is the decoder's current hidden state. The source hidden states weighted by the attention distribution form a context vector: $c_t = \sum_j \alpha_j h^j$. Next, the context vector and current hidden state are combined into an attention vector $a_t = \text{tanh}(W_a[c_t, h_t])$. Lastly, a softmax layer predicts the next token in the output sequence: $p(y^t_i|y^{<t}_i, x_i) = \text{softmax}(W_s a_t)$.

We add two auxiliary decoders when applying CVT. The auxiliary decoders share embedding and LSTM parameters with the primary decoder, but have different parameters for the attention mechanisms and softmax layers. For the first one, we restrict its view of the input by applying attention dropout, randomly zeroing out a fraction of its attention weights. The second one is trained to predict the next word in the target sequence rather than the current one:
$p_\theta^{\text{future}}(y^t_i|y^{<t}_i, x_i) = \text{softmax}(W_s^{\text{future}} a^{\text{future}}_{t - 1})$. 
Since there is no target sequence for unlabeled examples, we cannot apply teacher forcing to get an output distribution over the vocabulary from the primary decoder at each time step. Instead, we produce hard targets for the auxiliary modules by running the primary decoder with beam search on the input sequence. This idea has previously been applied to sequence-level knowledge distillation by \citet{Kim2016SequenceLevelKD} and makes the training procedure similar to back-translation \citep{Sennrich2016ImprovingNM}.

%-------------------------------------------------------
%                       Experiments
%-------------------------------------------------------

\section{Experiments}
We compare Cross-View Training against several strong baselines on seven tasks:

\vspace{1.5mm}\noindent \textbf{Combinatory Categorial Grammar (CCG) Supertagging:} 
We use data from CCGBank \citep{hockenmaier2007ccgbank}.

\vspace{1.5mm}\noindent \textbf{Text Chunking}: 
We use the CoNLL-2000 data \citep{tjong2000introduction}.

\vspace{1.5mm}\noindent \textbf{Named Entity Recognition (NER)}: 
We use the CoNLL-2003 data \citep{tjong2003introduction}.

\vspace{1.5mm}\noindent \textbf{Fine-Grained NER (FGN)}: 
We use the OntoNotes \citep{hovy2006ontonotes} dataset.

\vspace{1.5mm}\noindent \textbf{Part-of-Speech (POS) Tagging}: 
We use the Wall Street Journal portion of the Penn Treebank \citep{marcus1993building}.

\vspace{1.5mm}\noindent \textbf{Dependency Parsing:} 
We use the Penn Treebank converted to Stanford Dependencies version 3.3.0.

\vspace{1.5mm}\noindent \textbf{Machine Translation:} We use the English-Vietnamese translation dataset from IWSLT 2015 \citep{iwslt15}.
We report (tokenized) BLEU scores on the tst2013 test set. \\

We use the 1 Billion Word Language Model Benchmark \citep{chelba2013one} as a pool of unlabeled sentences for semi-supervised learning. 

\addtolength{\tabcolsep}{-2pt}
\begin{table*}[tb!]
\begin{tabularx}{\textwidth}{X | l l l l l | l l | l}
\ttop
  \multirow{2}{*}{Method} & CCG & Chunk & NER & FGN & POS & \multicolumn{2}{c |}{Dep. Parse} & Translate \\
             & Acc. & F1 & F1 & F1 & Acc. & UAS & LAS & BLEU \tsep
  Shortcut LSTM \cite{wu2017shortcut} & 95.1 &  &   &  & 97.53  & & & \Tstrut \\
  ID-CNN-CRF \cite{strubell2017fast} & &  & 90.7  & 86.8 &  & & & \\
  JMT$^\dagger$ \cite{hashimoto2016joint} & & 95.8 & & & 97.55 & 94.7 & 92.9 &  \\
  TagLM* \cite{peters2017semi} &  & 96.4  & 91.9  & & &  &  & \\
  ELMo* \cite{peters2018deep} &  &   & 92.2  & & & &  &  \tsep
  
  Biaffine \cite{Dozat2017Deep} &  &  & & & & 95.7 & 94.1  &  \Tstrut \\
  Stack Pointer \cite{ma2018stack} &  &  & & & & 95.9 & 94.2  &  \tsep
  
  Stanford \cite{Luong2015StanfordNM} &  & & & &  &  &  & 23.3 \Tstrut \\
  Google \cite{luong17} &  & & & &  &  &  & 26.1 \tsep
    
  Supervised & 94.9 & 95.1 & 91.2 & 87.5 & 97.60 & 95.1 & 93.3 & 28.9  \Tstrut \\
  Virtual Adversarial Training* & 95.1 & 95.1 & 91.8 & 87.9 & 97.64 & 95.4 & 93.7 & --  \\
  Word Dropout* & 95.2 & 95.8 & 92.1 & 88.1 & 97.66 & 95.6 & 93.8 & 29.3  \\
  ELMo (our implementation)* &  95.8 & 96.5  & 92.2  & 88.5 & 97.72 & 96.2 & 94.4  & 29.3  \\ 
  ELMo + Multi-task*$^\dagger$ & 95.9 &  96.8 & 92.3 & 88.4 & \textbf{97.79} & 96.4 & 94.8 & -- \\
  CVT* & 95.7 & 96.6 & 92.3  & 88.7 & 97.70 & 95.9 & 94.1 &  \textbf{29.6} \\
  CVT + Multi-task*$^\dagger$ & 96.0 &  96.9 & 92.4 & 88.4 & 97.76 & 96.4 &  94.8 & -- \\
  CVT + Multi-task + Large*$^\dagger$ & \textbf{96.1} &  \textbf{97.0} & \textbf{92.6} & \textbf{88.8} & 97.74 & \textbf{96.6} &  \textbf{95.0} & -- \tbottom
\end{tabularx}
\vspace{-1mm}
\caption{Results on the test sets. We report the mean score over 5 runs. Standard deviations in score are around 0.1 for NER, FGN, and  translation, 0.02 for POS, and 0.05 for the other tasks. See the appendix for results with them included. The +Large model has four times as many hidden units as the others, making it similar in size to the models when ELMo is included. 
* denotes semi-supervised and
$^\dagger$ denotes multi-task.} 
\vspace{-1mm}
\label{tab:nlp}
\end{table*}
\addtolength{\tabcolsep}{2pt}

\subsection{Model Details and Baselines}
We apply dropout during training, but not when running the primary prediction module to produce soft targets on unlabeled examples. In addition to the auxiliary prediction modules listed in Section~\ref{sec:models}, we find it slightly improves results to add another one that sees the whole input rather than a subset (but unlike the primary prediction module, does have dropout applied to its representations). 
Unless indicated otherwise, our models have LSTMs with 1024-sized hidden states and 512-sized projection layers. 
See the appendix for full training details and hyperparameters. We compare CVT with the following other semi-supervised learning algorithms:

\xhdr{Word Dropout} In this method, we only train the primary prediction module. When acting as a teacher it is run as normal, but when acting as a student, we  randomly replace some of the input words with a \texttt{REMOVED} token. This is similar to CVT in that it exposes the model to a restricted view of the input. However, it is less data efficient. By carefully designing the auxiliary prediction modules, it is possible to train the auxiliary prediction modules to match the primary one across many different views of the input a once, rather than just one view at a time.

\xhdr{Virtual Adversarial Training (VAT)} 
VAT \citep{miyato2015distributional} works like word dropout, but adds noise to the word embeddings of the student instead of dropping out words. Notably, the noise is chosen adversarially so it most changes the model's prediction. This method was applied successfully to semi-supervised text classification by \citet{miyato2016adversarial}. 
 
 \xhdr{ELMo} ELMo incorporates the representations from a large separately-trained language model into a task-specific model. Our implementaiton follows \citet{peters2018deep}. 
 When combining ELMo with multi-task learning, we allow each task to learn its own weights for the ELMo embeddings going into each prediction module. We found applying dropout to the ELMo embeddings was crucial for achieving good performance.

%-------------------------------------------------------
%                       Results
%-------------------------------------------------------

\subsection{Results}

\begin{figure}[t]
\includegraphics[width=0.47\textwidth]{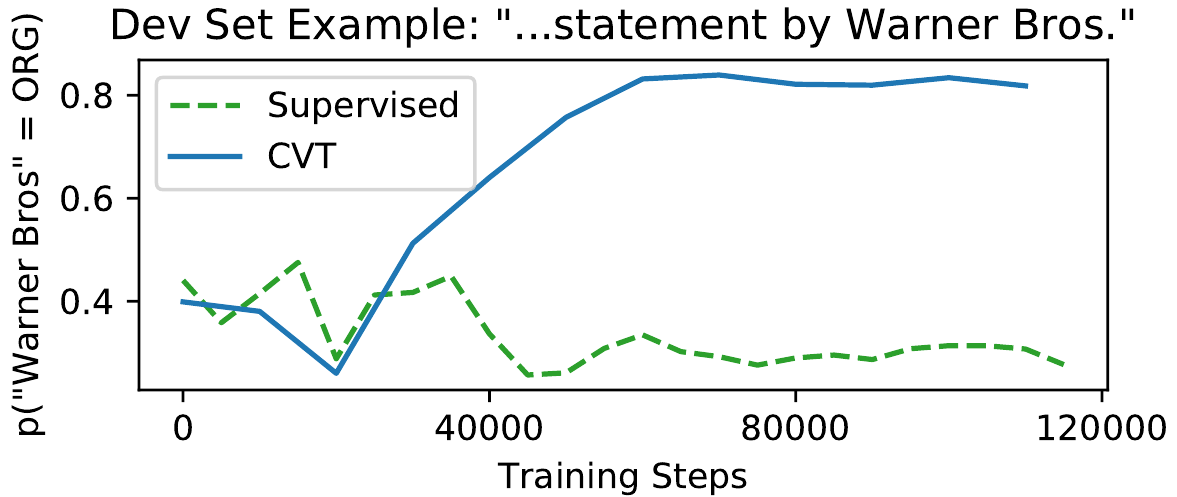}
\vspace{-3mm}
\caption{An NER example that CVT classifies correctly but supervised learning does not. ``Warner" only occurs as a last name in the train set, so the supervised model classifies ``Warner Bros" as a person. 
The CVT model also mistakenly classifies ``Warner Bros" as a person to start with, but as it sees more of the unlabeled data (in which ``Warner" occurs thousands of times) it learns that ``Warner Bros" is an organization.}
\label{fig:example}
\end{figure}

Results are shown in Table~\ref{tab:nlp}. 
CVT on its own outperforms or is comparable to the best previously published results on all tasks.
Figure~\ref{fig:example} shows an example win for CVT over supervised learning. .

Of the prior results listed in Table~\ref{tab:nlp}, only TagLM and ELMo are semi-supervised. These methods first train an enormous language model on unlabeled data and incorporate the representations produced by the language model into a supervised classifier.
Our base models use 1024 hidden units in their LSTMs (compared to 4096 in ELMo), require fewer training steps (around one pass over the billion-word benchmark rather than many passes), and do not require a pipelined training procedure.
Therefore, although they perform on par with ELMo, they are faster and simpler to train.
Increasing the size of our CVT+Multi-task model so it has 4096 units in its LSTMs like ELMo improves results further so they are significantly better than the ELMo+Multi-task ones.
We suspect there could be further gains from combining our method with language model pre-training, which we leave for future work.

\xhdr{CVT + Multi-Task}
We train a single shared-encoder CVT model to perform all of the tasks except machine translation (as it is quite different and requires more training time than the other ones). 
Multi-task learning improves results on all of the tasks except fine-grained NER, sometimes by large margins. 
Prior work on many-task NLP such as \citet{hashimoto2016joint} uses complicated architectures and training algorithms.
Our result shows that simple parameter sharing can be enough for effective many-task learning when the model is big and trained on a large amount of data. 

Interestingly, multi-task learning works better in conjunction with CVT than with ELMo.
We hypothesize that the ELMo models quickly fit to the data primarily using the ELMo vectors, which perhaps hinders the model from learning effective representations that transfer across tasks. 
We also believe CVT alleviates the danger of the model ``forgetting" one task while training on the other ones, a well-known problem in many-task learning \citep{Kirkpatrick2017OvercomingCF}.
During multi-task CVT, the model makes predictions about unlabeled examples across all tasks, creating (artificial) all-tasks-labeled examples, so the model does not only see one task at a time.
In fact, multi-task learning plus self training is similar to the Learning without Forgetting algorithm \citep{Li2016LearningWF}, which trains the model to keep its predictions on an old task unchanged when learning a new task. 
To test the value of all-tasks-labeled examples, we trained a multi-task CVT model that only computes $\lunsup$ on one task at a time (chosen randomly for each unlabeled minibatch) instead of for all tasks in parallel. The one-at-a-time model performs substantially worse (see Table~\ref{tab:oaat}).

\addtolength{\tabcolsep}{-3pt}
\begin{table}[h!]
\small
\begin{tabularx}{0.5\textwidth}{X l l l l l l}
\ttop 
Model        & CCG & Chnk & NER & FGN & POS & Dep. \Tstrut \tsep 
CVT-MT & 95.7 & 97.4 & 96.0 & 86.7 & 97.74 & 94.4 \Tstrut \\
\hspace{2mm}w/out all-labeled & 95.4 & 97.1 & 95.6 & 86.3 & 97.71 & 94.1  \tbottom
\end{tabularx}
\caption{Dev set performance of multi-task CVT with and without producing all-tasks-labeled examples.}
\label{tab:oaat}
\end{table}
\addtolength{\tabcolsep}{3pt}

\begin{figure}[t]
\includegraphics[width=0.5\textwidth]{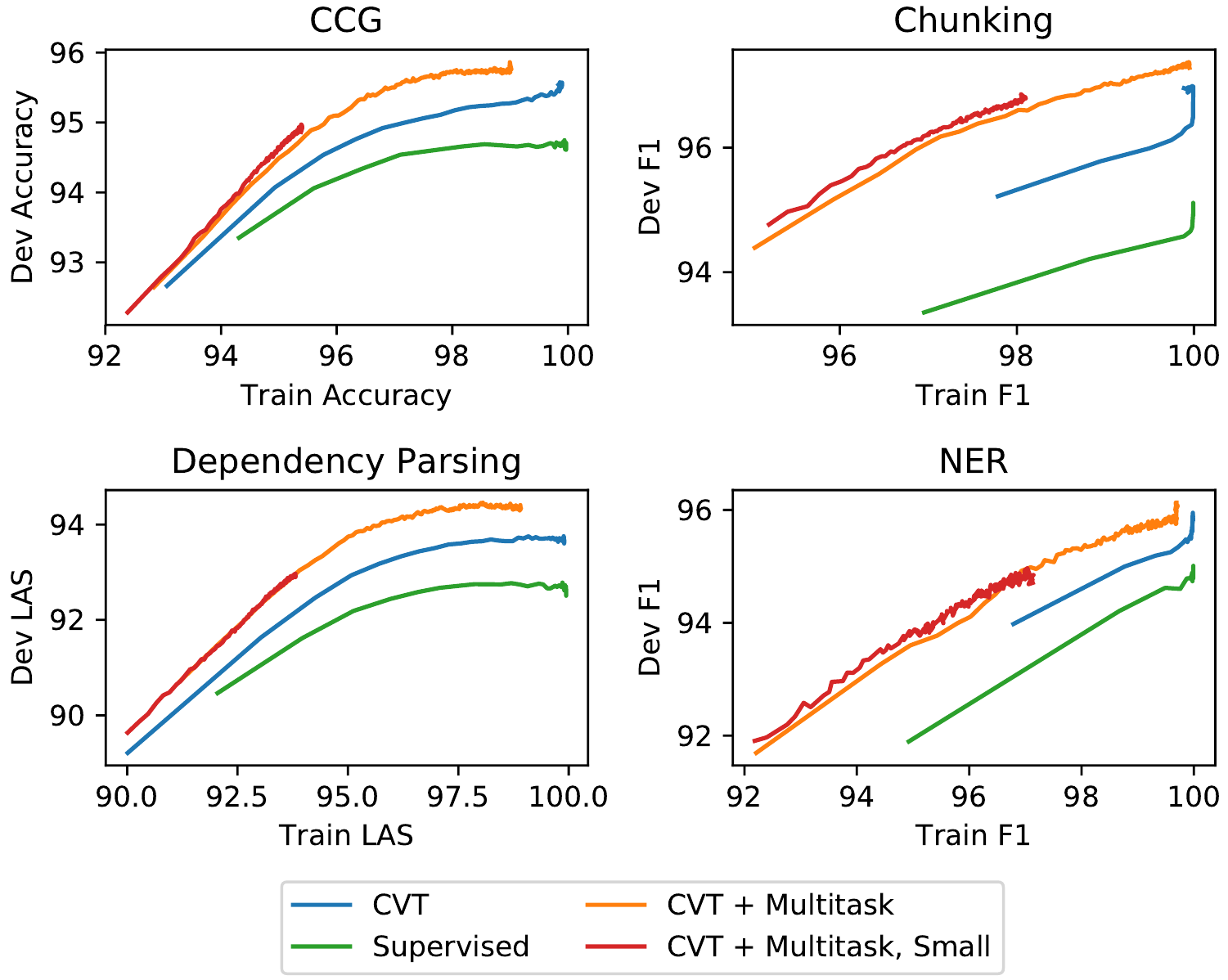}
\caption{Dev set vs.\ Train set accuracy for various methods. The ``small" model has 1/4 the LSTM hidden state size of the other ones (256 instead of 1024).}
\label{fig:curve}
\end{figure}

\xhdr{Model Generalization} 
In order to evaluate how our models generalize to the dev set from the train set, we plot the dev vs. train accuracy for our different methods as they learn (see Figure~\ref{fig:curve}). Both CVT and multi-task learning improve model generalization: for the same train accuracy, the models get better dev accuracy than purely supervised learning. 
Interestingly, CVT continues to improve in dev set accuracy while close to 100\% train accuracy for CCG, Chunking, and NER, perhaps because the model is still learning from unlabeled data even when it has completely fit to the train set. We also show results for a smaller multi-task + CVT model. Although it generalizes at least as well as the larger one, it halts making progress on the train set earlier. This suggests it is important to use sufficiently large neural networks for multi-task learning: otherwise the model does not have the capacity to fit to all the training data. 

\begin{figure*}[tb!]
\centering
\begin{minipage}{.46\textwidth}
\includegraphics[width=1.0\textwidth]{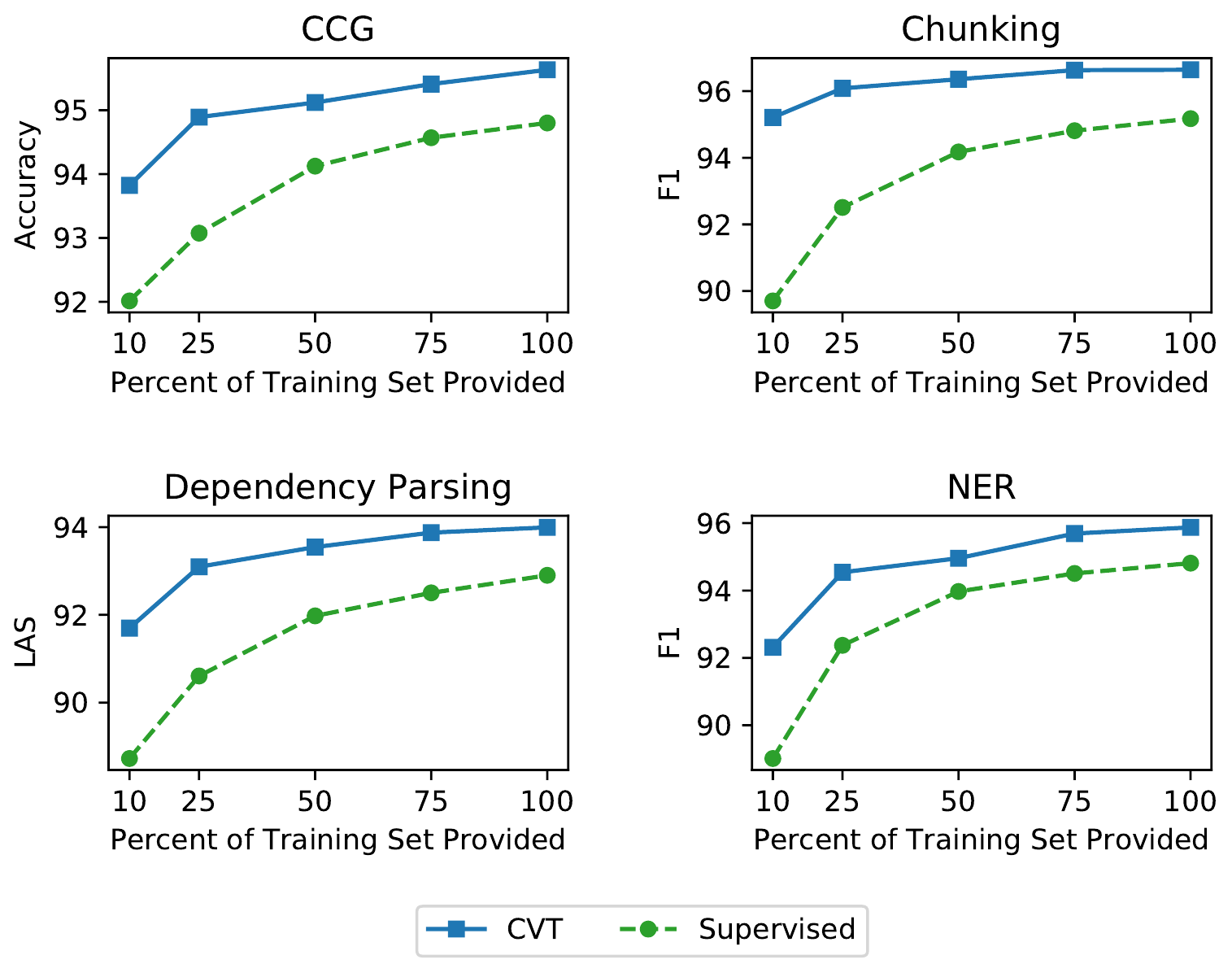}
\end{minipage} \hfill
\begin{minipage}{.46\textwidth}
\includegraphics[width=1.0\textwidth]{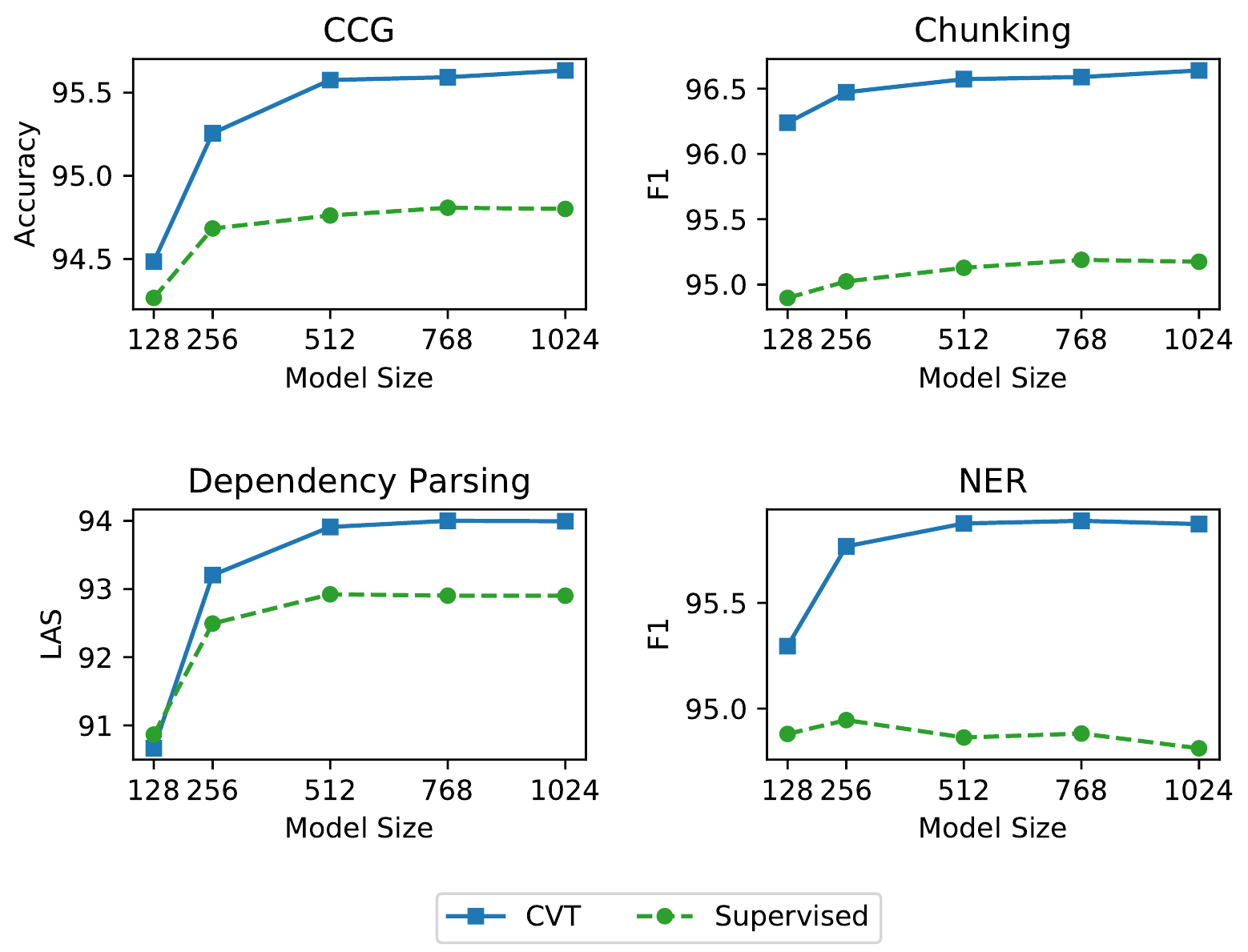}
\end{minipage}
\caption{Left: Dev set performance vs.\ percent of the training set provided to the model. 
Right: Dev set performance vs.\ model size. The $x$ axis shows the number of hidden units in the LSTM layers; the projection layers and other hidden layers in the network are half that size. 
Points correspond to the mean of three runs.}
\label{fig:size}
\end{figure*}

\xhdr{Auxiliary Prediction Module Ablation} We briefly explore which auxiliary prediction modules are more important for the sequence tagging tasks in Table~\ref{tab:abl}. 
We find that both kinds of auxiliary prediction modules improve performance, but that the future and past modules improve results more than the forward and backward ones, perhaps because they see a more restricted and challenging view of the input. 
\addtolength{\tabcolsep}{-1pt}
\begin{table}[ht!]
\small
\begin{tabularx}{0.5\textwidth}{X l l l l l }
\ttop 
Model        & CCG & Chnk & NER & FGN & POS \Tstrut \tsep 
Supervised     &  94.8 & 95.5 & 95.0 & 86.0 & 97.59  \Tstrut \\
CVT & 95.6 & 97.0 & 95.9 & 87.3 & 97.66  \\
\hspace{2mm}no fwd/bwd & --0.1 & --0.2 & --0.2  & --0.1 & --0.01  \\
\hspace{2mm}no future/past & --0.3 & --0.4 & --0.4 & --0.3 & --0.04 \tbottom
\end{tabularx}
\caption{Ablation study on auxiliary prediction modules for sequence tagging.}
\label{tab:abl}
\end{table}
\addtolength{\tabcolsep}{1pt}

\xhdr{Training Models on Small Datasets} We explore how CVT scales with dataset size by varying the amount of training data the model has access to.  
Unsurprisingly, the improvement of CVT over purely supervised learning grows larger as the amount of labeled data decreases (see Figure~\ref{fig:size}, left). Using only 25\% of the labeled data, our approach already performs as well or better than a fully supervised model using 100\% of the training data, demonstrating that CVT is particularly useful on small datasets.

\xhdr{Training Larger Models}
Most sequence taggers and dependency parsers in prior work use small LSTMs (hidden state sizes of around 300) because larger models yield little to no gains in performance \citep{reimers2017reporting}.
We found our own supervised approaches also do not benefit greatly from increasing the model size.
In contrast, when using CVT accuracy scales better with model size (see Figure~\ref{fig:size}, right).
This finding suggests the appropriate semi-supervised learning methods may enable the development of larger, more sophisticated models for NLP tasks with limited amounts of labeled data.

\xhdr{Generalizable Representations}
Lastly, we explore training the CVT+multi-task model on five tasks, freezing the encoder, and then only training a prediction module on the sixth task.
This tests whether the encoder's representations generalize to a new task not seen during its training.
Only training the prediction module is very fast because (1) the encoder (which is by far the slowest part of the model) has to be run over each example only once and (2) we do not back-propagate into the encoder.
Results are shown in Table~\ref{tab:gen}. 

\addtolength{\tabcolsep}{-3pt}
\begin{table}[h!]
\small
\begin{tabularx}{0.5\textwidth}{X l l l l l l}
\ttop 
Model        & CCG & Chnk & NER & FGN & POS & Dep. \Tstrut \tsep 
Supervised     &  94.8 & 95.6 & 95.0 & 86.0 & 97.59 & 92.9 \Tstrut \\
CVT-MT frozen & 95.1 & 96.6 & 94.6 & 83.2 & 97.66 & 92.5 \\
ELMo frozen & 94.3 & 92.2 & 91.3 & 80.6 & 97.50 & 89.4 \tbottom
\end{tabularx}
\caption{Comparison of single-task models on the dev sets. ``CVT-MT frozen" means we pretrain a CVT + multi-task model on five tasks, and then train only the prediction module for the sixth. ``ELMo frozen" means we train prediction modules (but no LSTMs) on top of ELMo embeddings.}
\label{tab:gen}
\end{table}
\addtolength{\tabcolsep}{3pt}

Training only a prediction module on top of multi-task representations works remarkably well, outperforming ELMo embeddings and sometimes even a vanilla supervised model, showing the multi-task model is building up effective representations for language.
In particular, the representations could be used like skip-thought vectors \citep{kiros2015skip} to quickly train models on new tasks without slow representation learning.

%-------------------------------------------------------
%                       Related Work
%-------------------------------------------------------

\section{Related Work}

\xhdr{Unsupervised Representation Learning}
Early approaches to deep semi-supervised learning pre-train neural models on unlabeled data, which has been successful for applications in computer vision \citep{jarrett2009best, lecun2010convolutional} and NLP\@.
Particularly noteworthy for NLP are algorithms for learning effective word embeddings \cite{collobert2011natural, Mikolov2013DistributedRO, pennington2014glove} and language model pretraining \citep{dai2015semi, ramachandran2016unsupervised, peters2018deep, howard2018universal, radford2018improving}.
Pre-training on other tasks such as machine translation has also been studied \citep{McCann2017LearnedIT}.
Other approaches train ``thought vectors" representing sentences through unsupervised \cite{kiros2015skip, Hill2016LearningDR} or supervised \cite{Conneau2017SupervisedLO} learning.

\xhdr{Self-Training}
One of the earliest approaches to semi-supervised learning is self-training \citep{scudder1965probability}, which has been successfully applied to NLP tasks such as word-sense disambiguation \citep{yarowsky1995unsupervised} and parsing \citep{mcclosky2006effective}. 
In each round of training, the classifier, acting as a ``teacher,'' labels some of the unlabeled data and adds it to the training set.
Then, acting as a ``student,'' it is retrained on the new training set.
Many recent approaches (including the consistentency regularization methods discussed below and our own method) train the student with soft targets from the teacher's output distribution rather than a hard label, making the procedure more akin to knowledge distillation \citep{hinton2015distilling}.
It is also possible to use multiple models or prediction modules for the teacher, such as in tri-training \citep{zhou2005tri, ruder2018strong}.

\xhdr{Consistency Regularization}
Recent works add noise (e.g., drawn from a Gaussian distribution) or apply stochastic transformations (e.g., horizontally flipping an image) to the student's inputs. 
This trains the model to give consistent predictions to nearby data points, encouraging distributional smoothness in the model.
Consistency regularization has been very successful for computer vision applications \citep{bachman2014learning,laine2016temporal,tarvainen2017weight}.
However, stochastic input alterations are more difficult to apply to discrete data like text, making consistency regularization less used for natural language processing. 
One solution is to add noise to the model's word embeddings \citep{miyato2016adversarial}; we compare against this approach in our experiments.
CVT is easily applicable to text because it does not require changing the student's inputs.

\xhdr{Multi-View Learning}
Multi-view learning on data where features can be separated into distinct subsets has been well studied \citep{Xu2013ASO}.
Particularly relevant are co-training \citep{blum1998combining} and co-regularization \citep{Sindhwani2005ACA}, which trains two models with disjoint views of the input.
On unlabeled data, each one acts as a ``teacher'' for the other model.
In contrast to these methods, our approach trains a single unified model where auxiliary prediction modules see different, but not necessarily independent views of the input.

\xhdr{Self Supervision}
Self-supervised learning methods train auxiliary prediction modules on tasks where performance can be measured without human-provided labels.
Recent work has jointly trained image classifiers with tasks like relative position and colorization \citep{doersch2017multi}, sequence taggers with language modeling \citep{rei2017semi}, and reinforcement learning agents with predicting changes in the environment \citep{jaderberg2016reinforcement}.
Unlike these approaches, our auxiliary losses are  based on self-labeling, not labels deterministically constructed from the input.

\xhdr{Multi-Task Learning}
There has been extensive prior work on multi-task learning \citep{ Caruana1997MultitaskL, Ruder2017AnOO}.  
For NLP, most work has focused on a small number of closely related tasks \citep{Luong2015MultitaskST,zhang2016stack, Sgaard2016DeepML, Peng2017DeepML}. 
Many-task systems are less commonly developed. 
\citet{Collobert2008AUA} propose a many-task system sharing word embeddings between the tasks, 
\citet{hashimoto2016joint} train a many-task model where the tasks are arranged hierarchically according to their linguistic level, and  
\citet{subramanian2018learning} train a shared-encoder many-task model for the purpose of learning better sentence representations for use in downstream tasks, not for improving results on the original tasks. 

%-------------------------------------------------------
%                       Conclusion
%-------------------------------------------------------

\section{Conclusion}
We propose Cross-View Training, a new method for semi-supervised learning.
Our approach allows models to effectively leverage their own predictions on unlabeled data, training them to produce effective representations that yield accurate predictions even when some of the input is not available. 
We achieve excellent results across seven NLP tasks, especially when CVT is combined with multi-task learning.

\section*{Acknowledgements}

We thank Abi See, Christopher Clark, He He, Peng Qi, Reid Pryzant, Yuaho Zhang,
and the anonymous reviewers for their thoughtful comments and suggestions.
We thank Takeru Miyato for help with his virtual adversarial training code and Emma Strubell for answering our questions about OntoNotes NER. 
Kevin is supported by a Google PhD Fellowship.

\bibliography{cvt}

\begin{thebibliography}{88}
\expandafter\ifx\csname natexlab\endcsname\relax\def\natexlab#1{#1}\fi

\bibitem[{Bachman et~al.(2014)Bachman, Alsharif, and
  Precup}]{bachman2014learning}
Philip Bachman, Ouais Alsharif, and Doina Precup. 2014.
\newblock Learning with pseudo-ensembles.
\newblock In \emph{NIPS}.

\bibitem[{Bahdanau et~al.(2015)Bahdanau, Cho, and Bengio}]{bahdanau2014neural}
Dzmitry Bahdanau, Kyunghyun Cho, and Yoshua Bengio. 2015.
\newblock Neural machine translation by jointly learning to align and
  translate.
\newblock In \emph{ICLR}.

\bibitem[{Blum and Mitchell(1998)}]{blum1998combining}
Avrim Blum and Tom Mitchell. 1998.
\newblock Combining labeled and unlabeled data with co-training.
\newblock In \emph{COLT}. ACM.

\bibitem[{Caruana(1997)}]{Caruana1997MultitaskL}
Rich Caruana. 1997.
\newblock Multitask learning.
\newblock \emph{Machine Learning}, 28:41--75.

\bibitem[{Cettolo et~al.(2015)Cettolo, Niehues, St{\"{u}}ker, Bentivogli,
  Cattoni, and Federico}]{iwslt15}
Mauro Cettolo, Jan Niehues, Sebastian St{\"{u}}ker, Luisa Bentivogli, Roldano
  Cattoni, and Marcello Federico. 2015.
\newblock The {IWSLT} 2015 evaluation campaign.
\newblock In \emph{International Workshop on Spoken Language Translation}.

\bibitem[{Chelba et~al.(2014)Chelba, Mikolov, Schuster, Ge, Brants, Koehn, and
  Robinson}]{chelba2013one}
Ciprian Chelba, Tomas Mikolov, Mike Schuster, Qi~Ge, Thorsten Brants, Phillipp
  Koehn, and Tony Robinson. 2014.
\newblock One billion word benchmark for measuring progress in statistical
  language modeling.
\newblock In \emph{INTERSPEECH}.

\bibitem[{Chiu and Nichols(2016)}]{chiu2015named}
Jason~PC Chiu and Eric Nichols. 2016.
\newblock Named entity recognition with bidirectional {LSTM-CNNs}.
\newblock \emph{Transactions of the Association for Computational Linguistics}.

\bibitem[{Choe and Charniak(2016)}]{choe2016parsing}
Do~Kook Choe and Eugene Charniak. 2016.
\newblock Parsing as language modeling.
\newblock In \emph{EMNLP}.

\bibitem[{Collobert and Weston(2008)}]{Collobert2008AUA}
Ronan Collobert and Jason Weston. 2008.
\newblock A unified architecture for natural language processing: deep neural
  networks with multitask learning.
\newblock In \emph{ICML}.

\bibitem[{Collobert et~al.(2011)Collobert, Weston, Bottou, Karlen, Kavukcuoglu,
  and Kuksa}]{collobert2011natural}
Ronan Collobert, Jason Weston, L{\'e}on Bottou, Michael Karlen, Koray
  Kavukcuoglu, and Pavel Kuksa. 2011.
\newblock Natural language processing (almost) from scratch.
\newblock \emph{Journal of Machine Learning Research}.

\bibitem[{Conneau et~al.(2017)Conneau, Kiela, Schwenk, Barrault, and
  Bordes}]{Conneau2017SupervisedLO}
Alexis Conneau, Douwe Kiela, Holger Schwenk, Lo{\"i}c Barrault, and Antoine
  Bordes. 2017.
\newblock Supervised learning of universal sentence representations from
  natural language inference data.
\newblock In \emph{EMNLP}.

\bibitem[{Dai and Le(2015)}]{dai2015semi}
Andrew~M Dai and Quoc~V Le. 2015.
\newblock Semi-supervised sequence learning.
\newblock In \emph{NIPS}.

\bibitem[{Dai et~al.(2017)Dai, Yang, Yang, Cohen, and
  Salakhutdinov}]{dai2017good}
Zihang Dai, Zhilin Yang, Fan Yang, William~W Cohen, and Ruslan Salakhutdinov.
  2017.
\newblock Good semi-supervised learning that requires a bad gan.
\newblock In \emph{NIPS}.

\bibitem[{Doersch and Zisserman(2017)}]{doersch2017multi}
Carl Doersch and Andrew Zisserman. 2017.
\newblock Multi-task self-supervised visual learning.
\newblock \emph{arXiv preprint arXiv:1708.07860}.

\bibitem[{Dozat and Manning(2017)}]{Dozat2017Deep}
Timothy Dozat and Christopher~D. Manning. 2017.
\newblock Deep biaffine attention for neural dependency parsing.
\newblock In \emph{ICLR}.

\bibitem[{Furlanello et~al.(2018)Furlanello, Lipton, Tschannen, Itti, and
  Anandkumar}]{furlanello2018born}
Tommaso Furlanello, Zachary~C Lipton, Michael Tschannen, Laurent Itti, and
  Anima Anandkumar. 2018.
\newblock Born again neural networks.
\newblock In \emph{ICML}.

\bibitem[{Graves and Schmidhuber(2005)}]{graves2005framewise}
Alex Graves and J{\"u}rgen Schmidhuber. 2005.
\newblock Framewise phoneme classification with bidirectional {LSTM} and other
  neural network architectures.
\newblock \emph{Neural Networks}, 18(5):602--610.

\bibitem[{Hashimoto et~al.(2017)Hashimoto, Xiong, Tsuruoka, and
  Socher}]{hashimoto2016joint}
Kazuma Hashimoto, Caiming Xiong, Yoshimasa Tsuruoka, and Richard Socher. 2017.
\newblock A joint many-task model: Growing a neural network for multiple nlp
  tasks.
\newblock In \emph{EMNLP}.

\bibitem[{Hill et~al.(2016)Hill, Cho, and Korhonen}]{Hill2016LearningDR}
Felix Hill, Kyunghyun Cho, and Anna Korhonen. 2016.
\newblock Learning distributed representations of sentences from unlabelled
  data.
\newblock In \emph{HLT-NAACL}.

\bibitem[{Hinton et~al.(2015)Hinton, Vinyals, and Dean}]{hinton2015distilling}
Geoffrey Hinton, Oriol Vinyals, and Jeff Dean. 2015.
\newblock Distilling the knowledge in a neural network.
\newblock \emph{arXiv preprint arXiv:1503.02531}.

\bibitem[{Hinton et~al.(2012)Hinton, Srivastava, Krizhevsky, Sutskever, and
  Salakhutdinov}]{hinton2012improving}
Geoffrey~E Hinton, Nitish Srivastava, Alex Krizhevsky, Ilya Sutskever, and
  Ruslan~R Salakhutdinov. 2012.
\newblock Improving neural networks by preventing co-adaptation of feature
  detectors.
\newblock \emph{arXiv preprint arXiv:1207.0580}.

\bibitem[{Hochreiter and Schmidhuber(1997)}]{hochreiter1997long}
Sepp Hochreiter and J{\"u}rgen Schmidhuber. 1997.
\newblock Long short-term memory.
\newblock \emph{Neural computation}, 9(8):1735--1780.

\bibitem[{Hockenmaier and Steedman(2007)}]{hockenmaier2007ccgbank}
Julia Hockenmaier and Mark Steedman. 2007.
\newblock {CCGbank}: a corpus of {CCG} derivations and dependency structures
  extracted from the {P}enn treebank.
\newblock \emph{Computational Linguistics}, 33(3):355--396.

\bibitem[{Hovy et~al.(2006)Hovy, Marcus, Palmer, Ramshaw, and
  Weischedel}]{hovy2006ontonotes}
Eduard Hovy, Mitchell Marcus, Martha Palmer, Lance Ramshaw, and Ralph
  Weischedel. 2006.
\newblock Ontonotes: the 90\% solution.
\newblock In \emph{HLT-NAACL}.

\bibitem[{Howard and Ruder(2018)}]{howard2018universal}
Jeremy Howard and Sebastian Ruder. 2018.
\newblock Universal language model fine-tuning for text classification.
\newblock In \emph{ACL}.

\bibitem[{Jaderberg et~al.(2017)Jaderberg, Mnih, Czarnecki, Schaul, Leibo,
  Silver, and Kavukcuoglu}]{jaderberg2016reinforcement}
Max Jaderberg, Volodymyr Mnih, Wojciech~Marian Czarnecki, Tom Schaul, Joel~Z
  Leibo, David Silver, and Koray Kavukcuoglu. 2017.
\newblock Reinforcement learning with unsupervised auxiliary tasks.
\newblock In \emph{ICLR}.

\bibitem[{Jarrett et~al.(2009)Jarrett, Kavukcuoglu, LeCun
  et~al.}]{jarrett2009best}
Kevin Jarrett, Koray Kavukcuoglu, Yann LeCun, et~al. 2009.
\newblock What is the best multi-stage architecture for object recognition?
\newblock In \emph{IEEE Conference on Computer Vision}.

\bibitem[{Kim and Rush(2016)}]{Kim2016SequenceLevelKD}
Yoon Kim and Alexander~M. Rush. 2016.
\newblock Sequence-level knowledge distillation.
\newblock In \emph{EMNLP}.

\bibitem[{Kirkpatrick et~al.(2017)Kirkpatrick, Pascanu, Rabinowitz, Veness,
  Desjardins, Rusu, Milan, Quan, Ramalho, Grabska-Barwinska, Hassabis, Clopath,
  Kumaran, and Hadsell}]{Kirkpatrick2017OvercomingCF}
James Kirkpatrick, Razvan Pascanu, Neil~C. Rabinowitz, Joel Veness, Guillaume
  Desjardins, Andrei~A. Rusu, Kieran Milan, John Quan, Tiago Ramalho, Agnieszka
  Grabska-Barwinska, Demis Hassabis, Claudia Clopath, Dharshan Kumaran, and
  Raia Hadsell. 2017.
\newblock Overcoming catastrophic forgetting in neural networks.
\newblock \emph{Proceedings of the National Academy of Sciences of the United
  States of America}, 114 13:3521--3526.

\bibitem[{Kiros et~al.(2015)Kiros, Zhu, Salakhutdinov, Zemel, Urtasun,
  Torralba, and Fidler}]{kiros2015skip}
Ryan Kiros, Yukun Zhu, Ruslan~R Salakhutdinov, Richard Zemel, Raquel Urtasun,
  Antonio Torralba, and Sanja Fidler. 2015.
\newblock Skip-thought vectors.
\newblock In \emph{Advances in neural information processing systems}, pages
  3294--3302.

\bibitem[{Krizhnevsky and Hinton(2009)}]{krizhevsky2009learning}
Alex Krizhnevsky and Geoffrey Hinton. 2009.
\newblock Learning multiple layers of features from tiny images.

\bibitem[{Kuncoro et~al.(2017)Kuncoro, Ballesteros, Kong, Dyer, Neubig, and
  Smith}]{kuncoro2017what}
Adhiguna Kuncoro, Miguel Ballesteros, Lingpeng Kong, Chris Dyer, Graham Neubig,
  and Noah~A. Smith. 2017.
\newblock What do recurrent neural network grammars learn about syntax?
\newblock In \emph{EACL}.

\bibitem[{Laine and Aila(2017)}]{laine2016temporal}
Samuli Laine and Timo Aila. 2017.
\newblock Temporal ensembling for semi-supervised learning.
\newblock In \emph{ICLR}.

\bibitem[{Lample et~al.(2016)Lample, Ballesteros, Subramanian, Kawakami, and
  Dyer}]{lample2016neural}
Guillaume Lample, Miguel Ballesteros, Sandeep Subramanian, Kazuya Kawakami, and
  Chris Dyer. 2016.
\newblock Neural architectures for named entity recognition.
\newblock In \emph{ACL}.

\bibitem[{LeCun et~al.(2010)LeCun, Kavukcuoglu, and
  Farabet}]{lecun2010convolutional}
Yann LeCun, Koray Kavukcuoglu, and Cl{\'e}ment Farabet. 2010.
\newblock Convolutional networks and applications in vision.
\newblock In \emph{ISCAS}. IEEE.

\bibitem[{Lewis et~al.(2016)Lewis, Lee, and Zettlemoyer}]{lewis2016lstm}
Mike Lewis, Kenton Lee, and Luke Zettlemoyer. 2016.
\newblock {LSTM CCG} parsing.
\newblock In \emph{HLT-NAACL}.

\bibitem[{Li and Hoiem(2016)}]{Li2016LearningWF}
Zhizhong Li and Derek Hoiem. 2016.
\newblock Learning without forgetting.
\newblock In \emph{ECCV}.

\bibitem[{Liu and Zhang(2017)}]{Liu2017in}
Jiangming Liu and Yue Zhang. 2017.
\newblock In-order transition-based constituent parsing.
\newblock \emph{TACL}.

\bibitem[{Liu et~al.(2017)Liu, Shang, Xu, Ren, Gui, Peng, and
  Han}]{liu2017empower}
Liyuan Liu, Jingbo Shang, Frank Xu, Xiang Ren, Huan Gui, Jian Peng, and Jiawei
  Han. 2017.
\newblock Empower sequence labeling with task-aware neural language model.
\newblock \emph{arXiv preprint arXiv:1709.04109}.

\bibitem[{Luong et~al.(2017)Luong, Brevdo, and Zhao}]{luong17}
Minh{-}Thang Luong, Eugene Brevdo, and Rui Zhao. 2017.
\newblock Neural machine translation (seq2seq) tutorial.
\newblock \emph{https://github.com/tensorflow/nmt}.

\bibitem[{Luong et~al.(2016)Luong, Le, Sutskever, Vinyals, and
  Kaiser}]{Luong2015MultitaskST}
Minh-Thang Luong, Quoc~V. Le, Ilya Sutskever, Oriol Vinyals, and Lukasz Kaiser.
  2016.
\newblock Multi-task sequence to sequence learning.
\newblock In \emph{ICLR}.

\bibitem[{Luong and Manning(2015)}]{Luong2015StanfordNM}
Minh-Thang Luong and Christopher~D. Manning. 2015.
\newblock Stanford neural machine translation systems for spoken language
  domains.
\newblock In \emph{IWSLT}.

\bibitem[{Luong et~al.(2015)Luong, Pham, and Manning}]{Luong2015EffectiveAT}
Thang Luong, Hieu Pham, and Christopher~D. Manning. 2015.
\newblock Effective approaches to attention-based neural machine translation.
\newblock In \emph{EMNLP}.

\bibitem[{Ma and Hovy(2016)}]{ma2016end}
Xuezhe Ma and Eduard Hovy. 2016.
\newblock End-to-end sequence labeling via bi-directional {LSTM-CNN-CRF}.
\newblock In \emph{ACL}.

\bibitem[{Ma and Hovy(2017)}]{ma2017neural}
Xuezhe Ma and Eduard Hovy. 2017.
\newblock Neural probabilistic model for non-projective mst parsing.
\newblock In \emph{IJCNLP}.

\bibitem[{Ma et~al.(2018)Ma, Hu, Liu, Peng, Neubig, and Hovy}]{ma2018stack}
Xuezhe Ma, Zecong Hu, Jingzhou Liu, Nanyun Peng, Graham Neubig, and Eduard
  Hovy. 2018.
\newblock Stack-pointer networks for dependency parsing.
\newblock In \emph{ACL}.

\bibitem[{Marcus et~al.(1993)Marcus, Marcinkiewicz, and
  Santorini}]{marcus1993building}
Mitchell~P Marcus, Mary~Ann Marcinkiewicz, and Beatrice Santorini. 1993.
\newblock Building a large annotated corpus of english: The {P}enn treebank.
\newblock \emph{Computational linguistics}, 19(2):313--330.

\bibitem[{McCann et~al.(2017)McCann, Bradbury, Xiong, and
  Socher}]{McCann2017LearnedIT}
Bryan McCann, James Bradbury, Caiming Xiong, and Richard Socher. 2017.
\newblock Learned in translation: Contextualized word vectors.
\newblock In \emph{NIPS}.

\bibitem[{McClosky et~al.(2006)McClosky, Charniak, and
  Johnson}]{mcclosky2006effective}
David McClosky, Eugene Charniak, and Mark Johnson. 2006.
\newblock Effective self-training for parsing.
\newblock In \emph{ACL}.

\bibitem[{Mikolov et~al.(2013)Mikolov, Sutskever, Chen, Corrado, and
  Dean}]{Mikolov2013DistributedRO}
Tomas Mikolov, Ilya Sutskever, Kai Chen, Gregory~S. Corrado, and Jeffrey Dean.
  2013.
\newblock Distributed representations of words and phrases and their
  compositionality.
\newblock In \emph{NIPS}.

\bibitem[{Miyato et~al.(2017{\natexlab{a}})Miyato, Dai, and
  Goodfellow}]{miyato2016adversarial}
Takeru Miyato, Andrew~M Dai, and Ian Goodfellow. 2017{\natexlab{a}}.
\newblock Adversarial training methods for semi-supervised text classification.
\newblock In \emph{ICLR}.

\bibitem[{Miyato et~al.(2017{\natexlab{b}})Miyato, Maeda, Koyama, and
  Ishii}]{miyato2017virtual}
Takeru Miyato, Shin-ichi Maeda, Masanori Koyama, and Shin Ishii.
  2017{\natexlab{b}}.
\newblock Virtual adversarial training: a regularization method for supervised
  and semi-supervised learning.
\newblock \emph{arXiv preprint arXiv:1704.03976}.

\bibitem[{Miyato et~al.(2016)Miyato, Maeda, Koyama, Nakae, and
  Ishii}]{miyato2015distributional}
Takeru Miyato, Shin-ichi Maeda, Masanori Koyama, Ken Nakae, and Shin Ishii.
  2016.
\newblock Distributional smoothing with virtual adversarial training.
\newblock In \emph{ICLR}.

\bibitem[{Park et~al.(2017)Park, Park, Shin, and Moon}]{park2017adversarial}
Sungrae Park, Jun-Keon Park, Su-Jin Shin, and Il-Chul Moon. 2017.
\newblock Adversarial dropout for supervised and semi-supervised learning.
\newblock \emph{arXiv preprint arXiv:1707.03631}.

\bibitem[{Peng et~al.(2017)Peng, Thomson, and Smith}]{Peng2017DeepML}
Hao Peng, Sam Thomson, and Noah~A. Smith. 2017.
\newblock Deep multitask learning for semantic dependency parsing.
\newblock In \emph{ACL}.

\bibitem[{Pennington et~al.(2014)Pennington, Socher, and
  Manning}]{pennington2014glove}
Jeffrey Pennington, Richard Socher, and Christopher Manning. 2014.
\newblock Glove: Global vectors for word representation.
\newblock In \emph{EMNLP}.

\bibitem[{Pereyra et~al.(2017)Pereyra, Tucker, Chorowski, Kaiser, and
  Hinton}]{pereyra2017regularizing}
Gabriel Pereyra, George Tucker, Jan Chorowski, {\L}ukasz Kaiser, and Geoffrey
  Hinton. 2017.
\newblock Regularizing neural networks by penalizing confident output
  distributions.
\newblock In \emph{ICLR}.

\bibitem[{Peters et~al.(2017)Peters, Ammar, Bhagavatula, and
  Power}]{peters2017semi}
Matthew~E Peters, Waleed Ammar, Chandra Bhagavatula, and Russell Power. 2017.
\newblock Semi-supervised sequence tagging with bidirectional language models.
\newblock In \emph{ACL}.

\bibitem[{Peters et~al.(2018)Peters, Neumann, Iyyer, Gardner, Clark, Lee, and
  Zettlemoyer}]{peters2018deep}
Matthew~E Peters, Mark Neumann, Mohit Iyyer, Matt Gardner, Christopher Clark,
  Kenton Lee, and Luke Zettlemoyer. 2018.
\newblock Deep contextualized word representations.
\newblock \emph{arXiv preprint arXiv:1802.05365}.

\bibitem[{Polyak(1964)}]{polyak1964some}
Boris~T Polyak. 1964.
\newblock Some methods of speeding up the convergence of iteration methods.
\newblock \emph{USSR Computational Mathematics and Mathematical Physics},
  4(5):1--17.

\bibitem[{Radford et~al.(2018)Radford, Narasimhan, Salimans, and
  Sutskever}]{radford2018improving}
Alec Radford, Karthik Narasimhan, Tim Salimans, and Ilya Sutskever. 2018.
\newblock Improving language understanding by generative pre-training.
\newblock \emph{https://blog.openai.com/language-unsupervised}.

\bibitem[{Ramachandran et~al.(2017)Ramachandran, Liu, and
  Le}]{ramachandran2016unsupervised}
Prajit Ramachandran, Peter~J Liu, and Quoc~V Le. 2017.
\newblock Unsupervised pretraining for sequence to sequence learning.
\newblock In \emph{EMNLP}.

\bibitem[{Rei(2017)}]{rei2017semi}
Marek Rei. 2017.
\newblock Semi-supervised multitask learning for sequence labeling.
\newblock In \emph{ACL}.

\bibitem[{Reimers and Gurevych(2017)}]{reimers2017reporting}
Nils Reimers and Iryna Gurevych. 2017.
\newblock Reporting score distributions makes a difference: Performance study
  of {LSTM}-networks for sequence tagging.
\newblock In \emph{EMNLP}.

\bibitem[{Ruder(2017)}]{Ruder2017AnOO}
Sebastian Ruder. 2017.
\newblock An overview of multi-task learning in deep neural networks.
\newblock \emph{arXiv preprint arXiv:1706.05098}.

\bibitem[{Ruder and Plank(2018)}]{ruder2018strong}
Sebastian Ruder and Barbara Plank. 2018.
\newblock Strong baselines for neural semi-supervised learning under domain
  shift.
\newblock In \emph{ACL}.

\bibitem[{Sajjadi et~al.(2016)Sajjadi, Javanmardi, and
  Tasdizen}]{sajjadi2016regularization}
Mehdi Sajjadi, Mehran Javanmardi, and Tolga Tasdizen. 2016.
\newblock Regularization with stochastic transformations and perturbations for
  deep semi-supervised learning.
\newblock In \emph{NIPS}.

\bibitem[{Salimans et~al.(2016)Salimans, Goodfellow, Zaremba, Cheung, Radford,
  and Chen}]{salimans2016improved}
Tim Salimans, Ian Goodfellow, Wojciech Zaremba, Vicki Cheung, Alec Radford, and
  Xi~Chen. 2016.
\newblock Improved techniques for training gans.
\newblock In \emph{NIPS}.

\bibitem[{Scudder(1965)}]{scudder1965probability}
H~Scudder. 1965.
\newblock Probability of error of some adaptive pattern-recognition machines.
\newblock \emph{IEEE Transactions on Information Theory}, 11(3):363--371.

\bibitem[{Sennrich et~al.(2016)Sennrich, Haddow, and
  Birch}]{Sennrich2016ImprovingNM}
Rico Sennrich, Barry Haddow, and Alexandra Birch. 2016.
\newblock Improving neural machine translation models with monolingual data.
\newblock In \emph{ACL}.

\bibitem[{Sindhwani and Belkin(2005)}]{Sindhwani2005ACA}
Vikas Sindhwani and Mikhail Belkin. 2005.
\newblock A co-regularization approach to semi-supervised learning with
  multiple views.
\newblock In \emph{ICML Workshop on Learning with Multiple Views}.

\bibitem[{S{\o}gaard and Goldberg(2016)}]{Sgaard2016DeepML}
Anders S{\o}gaard and Yoav Goldberg. 2016.
\newblock Deep multi-task learning with low level tasks supervised at lower
  layers.
\newblock In \emph{ACL}.

\bibitem[{Strubell et~al.(2017)Strubell, Verga, Belanger, and
  McCallum}]{strubell2017fast}
Emma Strubell, Patrick Verga, David Belanger, and Andrew McCallum. 2017.
\newblock Fast and accurate sequence labeling with iterated dilated
  convolutions.
\newblock In \emph{EMNLP}.

\bibitem[{Subramanian et~al.(2018)Subramanian, Trischler, Bengio, and
  Pal}]{subramanian2018learning}
Sandeep Subramanian, Adam Trischler, Yoshua Bengio, and Christopher~J Pal.
  2018.
\newblock Learning general purpose distributed sentence representations via
  large scale multi-task learning.
\newblock In \emph{ICLR}.

\bibitem[{Sutskever et~al.(2013)Sutskever, Martens, Dahl, and
  Hinton}]{sutskever2013importance}
Ilya Sutskever, James Martens, George Dahl, and Geoffrey Hinton. 2013.
\newblock On the importance of initialization and momentum in deep learning.
\newblock In \emph{ICML}.

\bibitem[{Sutskever et~al.(2014)Sutskever, Vinyals, and
  Le}]{sutskever2014sequence}
Ilya Sutskever, Oriol Vinyals, and Quoc~V Le. 2014.
\newblock Sequence to sequence learning with neural networks.
\newblock In \emph{NIPS}.

\bibitem[{Szegedy et~al.(2016)Szegedy, Vanhoucke, Ioffe, Shlens, and
  Wojna}]{szegedy2016rethinking}
Christian Szegedy, Vincent Vanhoucke, Sergey Ioffe, Jon Shlens, and Zbigniew
  Wojna. 2016.
\newblock Rethinking the inception architecture for computer vision.
\newblock In \emph{CVPR}.

\bibitem[{Tarvainen and Valpola(2017)}]{tarvainen2017weight}
Antti Tarvainen and Harri Valpola. 2017.
\newblock Weight-averaged consistency targets improve semi-supervised deep
  learning results.
\newblock In \emph{Workshop on Learning with Limited Labeled Data, NIPS}.

\bibitem[{Tjong Kim~Sang and Buchholz(2000)}]{tjong2000introduction}
Erik~F Tjong Kim~Sang and Sabine Buchholz. 2000.
\newblock Introduction to the {CoNLL}-2000 shared task: Chunking.
\newblock In \emph{CoNLL}.

\bibitem[{Tjong Kim~Sang and De~Meulder(2003)}]{tjong2003introduction}
Erik~F Tjong Kim~Sang and Fien De~Meulder. 2003.
\newblock Introduction to the {CoNLL}-2003 shared task: Language-independent
  named entity recognition.
\newblock In \emph{HLT-NAACL}.

\bibitem[{Verma et~al.(2018)Verma, Lamb, Beckham, Courville, Mitliagkis, and
  Bengio}]{verma2018manifold}
Vikas Verma, Alex Lamb, Christopher Beckham, Aaron Courville, Ioannis
  Mitliagkis, and Yoshua Bengio. 2018.
\newblock Manifold mixup: Encouraging meaningful on-manifold interpolation as a
  regularizer.
\newblock \emph{arXiv preprint arXiv:1806.05236}.

\bibitem[{Wei et~al.(2018)Wei, Liu, Wang, and Gong}]{wei2018improving}
Xiang Wei, Zixia Liu, Liqiang Wang, and Boqing Gong. 2018.
\newblock Improving the improved training of {W}asserstein {GAN}s.
\newblock In \emph{ICLR}.

\bibitem[{Wu et~al.(2017)Wu, Zhang, and Zong}]{wu2017shortcut}
Huijia Wu, Jiajun Zhang, and Chengqing Zong. 2017.
\newblock Shortcut sequence tagging.
\newblock \emph{arXiv preprint arXiv:1701.00576}.

\bibitem[{Xu et~al.(2013)Xu, Tao, and Xu}]{Xu2013ASO}
Chang Xu, Dacheng Tao, and Chao Xu. 2013.
\newblock A survey on multi-view learning.
\newblock \emph{arXiv preprint arXiv:1304.5634}.

\bibitem[{Yarowsky(1995)}]{yarowsky1995unsupervised}
David Yarowsky. 1995.
\newblock Unsupervised word sense disambiguation rivaling supervised methods.
\newblock In \emph{ACL}.

\bibitem[{Zhang et~al.(2018)Zhang, Cisse, Dauphin, and
  Lopez-Paz}]{zhang2017mixup}
Hongyi Zhang, Moustapha Cisse, Yann~N Dauphin, and David Lopez-Paz. 2018.
\newblock mixup: Beyond empirical risk minimization.
\newblock In \emph{ICLR}.

\bibitem[{Zhang and Weiss(2016)}]{zhang2016stack}
Yuan Zhang and David Weiss. 2016.
\newblock Stack-propagation: Improved representation learning for syntax.
\newblock In \emph{ACL}.

\bibitem[{Zhou and Li(2005)}]{zhou2005tri}
Zhi-Hua Zhou and Ming Li. 2005.
\newblock Tri-training: Exploiting unlabeled data using three classifiers.
\newblock \emph{IEEE Transactions on knowledge and Data Engineering}.

\end{thebibliography}
\bibliographystyle{acl_natbib_nourl}

\appendix
\section{Detailed Results} We provide a more detailed version of the test set results in the paper, adding two decimals of precision, standard deviations of the 5 runs for each model, and more prior work, in Table~\ref{tab:detailed-nlp}.

\section{Model Details}
Our models use two layer CNN-BiLSTM encoders \citep{chiu2015named,ma2016end,lample2016neural} and task-specific prediction modules.
See Section~\ref{sec:models} of the paper for details.
We provide a few minor details not covered there below.

\xhdr{Sequence Tagging} 
For Chunking and Named Entity Recognition, we use a BIOES tagging scheme. We apply label smoothing \citep{szegedy2016rethinking, pereyra2017regularizing} with a rate of 0.1 to the target labels when training on the labeled data.

\xhdr{Dependency Parsing}
We omit punctuation from evaluation, which is standard practice for the PTB-SD 3.3.0 dataset. ROOT is represented with a fixed vector $h_{\texttt{ROOT}}$ instead of using a vector from the encoder, but otherwise  dependencies coming from ROOT are scored the same way as the other dependencies.  

\xhdr{Machine Translation}
We apply dropout to the output of each LSTM layer in the decoder. Our implementation is heavily based off of the Google NMT Tutorial\footnote{\url{https://github.com/tensorflow/nmt}} \citep{luong17}. We attribute our significantly better results to using pre-trained word embeddings, a character-level CNN, a larger model, stronger regularization, and better hyperparameter tuning. Target words occurring 5 or fewer times in the train set are replaced with a \texttt{UNK} token (but not during evaluation). We use a beam size of 10 when performing beam search. 
We found it slightly beneficial to apply label smoothing with a rate of 0.1 to the teacher's predictions (unlike our other tasks, the teacher only provides hard targets to the students for translation).

\xhdr{Multi-Task Learning}
Several of our datasets are constructed from the Penn Treebank. 
However, we treat them as separate rather than providing examples labeled across multiple tasks to our model during supervised training.
Furthermore, the Penn Treebank tasks do not all use the same train/dev/test splits. We ensure the training split of one task never overlaps the evaluation split of another by discarding the overlapping examples from the train sets. 

\xhdr{Other Details}
We apply dropout \citep{hinton2012improving} to the word embeddings and outputs of each Bi-LSTM. We use an exponential-moving-average (EMA) of the model weights from training for the final model; we found this to slightly improve accuracy and significantly reduce the variance in accuracy between models trained with different random initializations. The model is trained using SGD with momentum \citep{polyak1964some, sutskever2013importance}. Word embeddings are initialized with GloVe vectors \citep{pennington2014glove} and fine-tuned during training. The full set of model hyperparameters are listed in Table~\ref{tab:hyper}.

\xhdr{Baselines}
Baselines were run with the same architecture and hyperparameters as the CVT model. For the ``word dropout" model, we randomly replace words in the input sentence with a \texttt{REMOVED} token with probability 0.1 (this value worked well on the dev sets). For Virtual Adversarial Training, we set the norm of the perturbation to be 1.5 for CCG, 1.0 for Dependency Parsing, and 0.5 for the other tasks (these values worked best on the dev sets). Otherwise, the implementation is as described in \citep{miyato2016adversarial}; we based our implementation off of their code\footnote{\url{https://github.com/tensorflow/models/tree/master/research/adversarial_text}}. We were unable to successfully apply VAT to machine translation, perhaps because the student is provided hard targets for that task. 
For ELMo, we applied dropout to the ELMo embeddings before they are incorporated into the rest of the model.
When training the multi-task ELMo model, each prediction module has its own set of softmax-normalized weights ($s_j^{task}$ in \citep{peters2018deep}) for the ELMo emeddings going into the task-specific prediction modules. All tasks share the same $s_j$ weights for the ELMo embeddings going into the shared Bi-LSTM encoder.

\addtolength{\tabcolsep}{-1pt}
\begin{table*}[b!]
\resizebox{1.0\textwidth}{!}{
\begin{tabular}{l | l l l l l | l l | l }
\ttop
 \multirow{2}{*}{Method} & CCG & Chunking & NER  & FGN & POS & \multicolumn{2}{c |}{Dependency Parsing} & Translation  \\
             & Acc. & F1 & F1 & F1 & Acc. & UAS & LAS & BLEU \tsep
             
 LSTM-CNN-CRF \cite{ma2016end} &  & & 91.21 &  & 97.55  &  & \Tstrut\\
 LSTM-CNN \cite{chiu2015named} &  & & 91.62 $\pm$ 0.33 &  86.28 $\pm$ 0.26 & &  &  \\
 ID-CNN-CRF \cite{strubell2017fast} &  & & 90.65 $\pm$ 0.15 &  86.84 $\pm$ 0.19 & &  & \\
 Tri-Trained LSTM \cite{lewis2016lstm} & 94.7 &  &  & &  &  & &  \\
 Shortcut LSTM \cite{wu2017shortcut} & 95.08 &  &  & & 97.53 & & &  \\
  JMT* \cite{hashimoto2016joint} &  & 95.77 &  & & 97.55 & 94.67 & 92.90 & \\
 LM-LSTM-CNN-CRF \cite{liu2017empower} &  & 95.96 $\pm$ 0.08 & 91.71 $\pm$ 0.10  & & 97.53 $\pm$ 0.03 & & &  \\
 TagLM$^\dagger$ \cite{peters2017semi} &  & 96.37 $\pm$ 0.05  & 91.93 $\pm$ 0.19  &  & & & & \\
 ELMo$^\dagger$ \cite{peters2018deep} &  &   & 92.22 $\pm$ 0.10  & & & & &  \tsep
 NPM \cite{ma2017neural}  &  &  & & & & 94.9 & 93.0 &  \Tstrut \\
Deep Biaffine \cite{Dozat2017Deep} &  & & &  & & 95.74 & 94.08 &  \\ 
Stack Pointer \cite{ma2018stack} &  & & &  & & 95.87 & 94.19 &  \tsep
Stanford \cite{Luong2015StanfordNM} &  & & &  & &  &  & 23.3  \Tstrut \\ 
Google \cite{luong17} &  & & &  & & &  &  26.1 \tsep
Supervised & 94.94 $\pm$ 0.02 & 95.10 $\pm$ 0.06 & 91.16 $\pm$ 0.09 & 87.48 $\pm$ 0.08 & 97.60 $\pm$ 0.02 & 95.08 $\pm$ 0.03 & 93.27 $\pm$ 0.03 & 28.88 $\pm$ 0.12 \Tstrut \\
Virtual Adversarial Training* & 95.07 $\pm$ 0.04 & 95.06 $\pm$ 0.06 & 91.75 $\pm$ 0.10 & 87.91 $\pm$ 0.11 & 97.64 $\pm$ 0.03 & 95.44 $\pm$ 0.06 & 93.72 $\pm$ 0.07 & --    \\
Word Dropout* & 95.20 $\pm$ 0.04 & 95.79 $\pm$ 0.08 & 92.14 $\pm$ 0.11 & 88.06 $\pm$ 0.09 & 97.66 $\pm$ 0.01 & 95.56 $\pm$ 0.05 & 93.80 $\pm$ 0.08 & 29.33 $\pm$ 0.10    \\
ELMo* & 95.79 $\pm$ 0.04 & 96.50 $\pm$ 0.03 & 92.24 $\pm$ 0.09 & 88.49 $\pm$ 0.12 & 97.72 $\pm$ 0.01 & 96.22 $\pm$ 0.05 & 94.44 $\pm$ 0.06 & 29.34 $\pm$ 0.11 \\
ELMo + Multi-task*$^\dagger$ & 95.91 $\pm$ 0.05 & 96.83 $\pm$ 0.03 & 92.32 $\pm$ 0.12 & 88.37 $\pm$ 0.16 & \textbf{97.79 $\pm$ 0.03} & 96.40 $\pm$ 0.04 & 94.79 $\pm$ 0.05 & -- \\
CVT* & 95.65 $\pm$ 0.04 & 96.58 $\pm$ 0.04 & 92.34 $\pm$ 0.06 & 88.68 $\pm$ 0.14 & 97.70 $\pm$ 0.03 & 95.86 $\pm$ 0.03 & 94.06 $\pm$ 0.02 & \textbf{29.58 $\pm$ 0.07}   \\
CVT + Multi-Task*$^\dagger$ & 95.97 $\pm$ 0.04 &  96.85 $\pm$ 0.05 &  92.42 $\pm$ 0.08 & 88.42 $\pm$ 0.13 & 97.76 $\pm$ 0.02 &  96.44 $\pm$ 0.04 &  94.83 $\pm$ 0.06 & -- \\
CVT + Multi-Task + Large*$^\dagger$ & \textbf{96.05 $\pm$ 0.03} &  \textbf{96.98 $\pm$ 0.05} &  \textbf{92.61 $\pm$ 0.09} & \textbf{88.81 $\pm$ 0.09} & 97.74 $\pm$ 0.02 &  \textbf{96.61 $\pm$ 0.04} &  \textbf{95.02 $\pm$ 0.04} & -- 
\tbottom
\end{tabular}
}
\vspace{0mm} \\ 
\caption{Results on the test sets for all tasks. We report the means and standard deviations of 5 runs. 
The +Larger model has four times as many hidden units as the others, making it similar in size to the models when ELMo is included. 
For dependency parsing, we omit results from \citet{choe2016parsing}, \citet{kuncoro2017what}, and \citet{Liu2017in} because these train constituency parsers and convert the system outputs to dependency parses. They produce higher scores, but have access to more information during training and do not apply to datasets without constituency annotations. * denotes semi-supervised and $^\dagger$ denotes multi-task.}
\vspace{-1mm}
\label{tab:detailed-nlp}
\end{table*}
\addtolength{\tabcolsep}{1pt}

\begin{table*}[t!]
\begin{tabular}{l l}
\ttop
Parameter & Value\Tstrut\tsep
Word Embeddings Initializiation & 300d GloVe 6B \Tstrut\\
Character Embedding Size & 50 \\
Character CNN Filter Widths & [2, 3, 4] \\
Character CNN Num Filters & 300 (100 per filter width) \\
Encoder LSTM sizes & 1024 for the first layer, 512 for the second one \\
Encoder LSTM sizes, ``Large" model & 4096 for the first layer, 2048 for the second one \\
LSTM projection layer size & 512 \\
Hidden layer sizes & 512  \\
Dropout & 0.5 for labeled examples, 0.8 for unlabeled examples\\
EMA coefficient & 0.998\\
Learning rate & $0.5 / (1 + 0.005t^{0.5})$ ($t$ is number of SGD updates so far)\\
Momentum & 0.9 \\
Batch size & 64 sentences\tbottom
\end{tabular}
\caption{Hyperparameters for the model.}
\label{tab:hyper}
\end{table*}

\section{CVT for Image Recognition}
\label{sec:img}

Although the focus of our work is on NLP, we also applied CVT to image recognition and found it performs competitively with existing methods. Most of the semi-supervised image recognition approaches we compare against rely on the inputs being continuous, so they would be difficult to apply to text. More specifically, consistency regularization methods \citep{sajjadi2016regularization, laine2016temporal, miyato2017virtual} rely on adding continuous noise and applying image-specific transformations like cropping to inputs, GANs \citep{salimans2016improved, wei2018improving} are very difficult to train on text due to its discrete nature, and mixup \citep{zhang2017mixup, verma2018manifold} requires a way of smoothly interpolating between different inputs. 

\xhdr{Approach}
Our image recognition models are based on Convolutional Neural Networks, which produce a set of features $H(x_i) \in \mathbb{R}^{n \times n \times d}$ from an image $x_i$.
The first two dimensions of $H$ index into the spatial coordinates of feature vectors and $d$ is the size of the feature vectors.
For shallower CNNs, a particular feature vector corresponds to a region of the input image.
For example, $H_{0, 0}$ would be a $d$-dimensional vector of features extracted from the upper left corner.
For deeper CNNs, a particular feature vector would be extracted from the whole image, but still only use a ``region'' of the representations from an earlier layer.
The CNNs in our experiments are all in the first category.

The primary prediction layers of our CNNs take as input the mean of $H$ over the first two dimensions, which results in a $d$-dimensional vector that is fed into a softmax layer:
\[
\resizebox{0.49\textwidth}{!}{
$\model{x_i} = \text{softmax}(W\texttt{global\_average\_pool}(H) + b)$
}
\]

We add $n^2$ auxiliary prediction layers to the top of the CNN. The $j$th layer takes a single feature vector as input:
\[
\modelj{x_i} = \text{softmax}(W^j H_{\lfloor j / n \rfloor, j \text{ mod } n} + b_j)
\]

\xhdr{Data}
We evaluated our models on the CIFAR-10 \citep{krizhevsky2009learning} dataset. 
Following previous work, we make the datasets semi-supervised by only using the provided labels for a subset of the examples in the training set; the rest are treated as unlabeled examples.

\xhdr{Model} We use the convolutional neural network from \citet{miyato2017virtual}, adapting their TensorFlow implementation\footnote{\url{https://github.com/takerum/vat_tf}}.
Their model contains 9 convolutional layers and 2 max pooling layers. See Appendix D of Miyato et al.'s paper for more details.

We add 36 auxiliary softmax layers to the $6 \times 6$ collection of feature vectors produced by the CNN.
Each auxiliary layer sees a patch of the image ranging in size from $21 \times 21$ pixels (the corner) to $29 \times 29$ pixels (the center) of the $32 \times 32$ pixel images.
For some experiments, we combine CVT with standard consistency regularization by adding a perturbation (e.g., a small random vector) to the student's inputs when computing $\lunsup$.

\addtolength{\tabcolsep}{15pt}
\begin{table*}[h!]
\begin{tabular}{l l l}
\ttop
\multirow{2}{*}{Method} & CIFAR-10 & CIFAR-10+\Tstrut \\
             &  \multicolumn{2}{c}{$4000$ labels}\tsep
 GAN \cite{salimans2016improved} & -- & 18.63 $\pm$ 2.32\Tstrut\\
 Stochastic Transformations \cite{sajjadi2016regularization}  & -- & 11.29 $\pm$ 0.24\\
 $\Pi$ model \cite{laine2016temporal}  & 16.55 $\pm$ 0.29 & 12.36 $\pm$ 0.31 \\
 Temporal Ensemble \cite{laine2016temporal}  & -- & 12.16 $\pm$ 0.24 \\
 Mean Teacher \cite{tarvainen2017weight} & -- & 12.31 $\pm$ 0.28 \\
 Complement GAN \cite{dai2017good}  & 14.41 $\pm$ 0.30 & -- \\
 VAT \cite{miyato2017virtual} & 13.15 & 10.55 \\
 VAdD \cite{park2017adversarial} & -- & 11.68 $\pm$ 0.19 \\
 VAdD + VAT \cite{park2017adversarial}  & -- & \textbf{10.07 $\pm$ 0.11} \\
 SNGT + $\Pi$ model \cite{luong17} & 13.62 $\pm$ 0.17 & 11.00 $\pm$ 0.36 \\
 SNGT + VAT \cite{luong17}  & \textbf{12.49 $\pm$ 0.36} & \textbf{9.89 $\pm$ 0.34} \\
 Consistency + WGAN \cite{wei2018improving} & -- & \textbf{9.98 $\pm$ 0.21} \\
 Manifold Mixup \cite{verma2018manifold} & -- &  \textbf{10.26 $\pm$ 0.32} 
 \tsep
 Supervised & 23.61 $\pm$ 0.60 & 19.61 $\pm$ 0.56 \Tstrut\\
 VAT (ours)  & 13.29 $\pm$ 0.33 & 10.90 $\pm$ 0.31\\
 CVT, no input perturbation  & 14.63 $\pm$ 0.20 & 12.44 $\pm$ 0.27 \\
 CVT, random input perturbation & 13.80 $\pm$ 0.30 & 11.10 $\pm$ 0.26 \\
 CVT, adversarial input perturbation  & \textbf{12.01 $\pm$ 0.11} & \textbf{10.11 $\pm$ 0.15}\tbottom
\end{tabular}
\vspace{1mm}
\caption{
		Error rates on semi-supervised CIFAR-10. We report means and standard deviations from 5 runs.
		CIFAR-10+ refers to results where data augmentation (random translations of the input image) was applied. For some of our models we add a random or adversarially chosen perturbation to the student model's inputs, which is done in most consistency regularization methods.
}
\label{tab:vision}
\end{table*}
\addtolength{\tabcolsep}{-15pt}

\xhdr{Results}
The results are shown in Table~\ref{tab:vision}. Unsurprisingly, adding continuous noise to the inputs works much better with images, where the inputs are naturally continuous, than with language. Therefore we see much better results from VAT on semi-supervised CIFAR-10 compared to on our NLP tasks. 
However, we still find incorporating CVT improves over models without CVT. Our CVT + VAT models are competitive with current start-of-the-art approaches.
We found the gains from CVT are larger when no data augmentation is applied, perhaps because random translations of the input expose the model to different ``views'' in a similar manner as with CVT.

\section{Negative Results}
We briefly describe a few ideas we implemented that did not seem to be effective in initial experiments. Note these findings are from early one-off experiments. We did not pursue them further after our first attempts did not pan out, so it is possible that some of these approaches could be effective with the proper adjustments and tuning. 
\begin{itemize}
    \item \textbf{Hard vs soft targets:} Classic self-training algorithms train the student model with one-hot ``hard" targets corresponding to the teacher's highest probability prediction. In our experiments, this decreased performance compared to using soft targets. This finding is consistent with research on knowledge distillation \citep{hinton2015distilling, furlanello2018born} where soft targets also work notably better than hard targets.  
    \item \textbf{Confidence thresholding:} Classic self-training often only trains the student on a subset of the unlabeled examples on which the teacher has confident predictions (i.e., the output distribution has low entropy). We tried both ``hard" (where the student ignores low-confidence examples) and ``soft" (where examples are weighted according to the teacher's confidence) versions of this for training our models, but they did not seem to improve performance. 
    \item \textbf{Mean Teacher:} The Mean Teacher method \citep{tarvainen2017weight} tracks an exponential moving average (EMA) of model weights, which are used to produce targets for the students. The idea is that these targets may be better quality due to a self-ensembling effect. However, we found this approach to have little to no benefit in our experiments, although using EMA model weights at test time did improve results slightly.
    \item \textbf{Purely supervised CVT:} Lastly, we explored adding cross-view losses to purely supervised classifiers. We hoped that adding auxiliary softmax layers with different views of the input would act as a regularizer on the model. However, we found little to no benefit from this approach. This negative result suggests that the gains from CVT are from the improved semi-supervised learning mechanism, not the additional prediction layers regularizing the model.
\end{itemize}

\end{document}